\pdfoutput=1

\documentclass[11pt]{article}
\usepackage[dvipsnames]{xcolor}

\usepackage{acl}

\usepackage{times}
\usepackage{latexsym}

\usepackage[T1]{fontenc}

\usepackage[utf8]{inputenc}

\usepackage{microtype}
\usepackage{url}
\usepackage{graphicx}
\usepackage{wrapfig}
\usepackage{amsmath}
\usepackage{amssymb}
\usepackage{color,xcolor,colortbl}
\usepackage{enumitem}
\usepackage{bm,bbm}
\usepackage{booktabs}
\usepackage{mathtools}
\usepackage{array}
\usepackage{subcaption}
\usepackage{pbox}
\usepackage{pifont}
\usepackage{multirow}
\usepackage[scaled=0.7]{beramono}
\usepackage{tabularx}
\usepackage{tablefootnote}
\usepackage{float}
\usepackage{booktabs}
\usepackage{arydshln}
\newcommand{\cmark}{\ding{51}}%
\newcommand{\xmark}{\ding{55}}%
\title{NoteChat: A Dataset of Synthetic Patient-Physician Conversations Conditioned on Clinical Notes}

\author{Junda Wang \thanks{* indicates equal contribution} $^1$, Zonghai Yao\footnotemark[1] $^1$, \bf{Zhichao Yang}$^1$, \bf{Huixue Zhou}$^2$, \bf{Rumeng Li}$^1$ \\ \bf{Xun Wang}$^3$, \bf{Yucheng Xu}$^4$, \bf{Hong Yu}$^{1, 5}$\\
University of Massachusetts, Amherst$^1$, University of Minnesota$^2$, Microsoft$^3$ \\ University of Edinburgh$^4$, University of Massachusetts, Lowell$^5$\\
{\tt \{jundawang, zonghaiyao, zhichaoyang, hongyu\}@umass.edu}\\ 
}

\begin{document}

\maketitle
 \begin{abstract}
We introduce NoteChat, a novel cooperative multi-agent framework leveraging Large Language Models (LLMs) to generate patient-physician dialogues. NoteChat embodies the principle that an ensemble of role-specific LLMs, through structured role-play and strategic prompting, can perform their assigned roles more effectively. The synergy among these role-playing LLMs results in a cohesive and efficient dialogue generation. Evaluation on MTS-dialogue~\cite{abacha2023empirical, mediqa-chat-2023}, a benchmark dataset for patient-physician dialogues-note pairs, shows that models trained with the augmented synthetic patient-physician dialogues by NoteChat~\footnote{
Our synthetic patient-physician dialogue data is in supplementary material and are publicly available together with all codes and prompts at github:https://github.com/believewhat/Dr.NoteAid.
} outperforms other state-of-the-art models for generating clinical notes. Our comprehensive automatic and human evaluation demonstrates that NoteChat substantially surpasses state-of-the-art models like ChatGPT and GPT-4 up to 22.78\% by domain experts in generating superior synthetic patient-physician dialogues based on clinical notes. NoteChat has the potential to engage patients directly and help clinical documentation, a leading cause of physician burnout~\cite{budd2023burnout}. 

\end{abstract}

\section{Introduction}

Clinical dialogue is an essential part of clinical workflow. Clinical documentation is a two-step process. It first engages patients through conversation to collect patient-specific information such as demographic information, family history of diseases, and signs and symptoms and then generates electronic health records (EHRs) from the dialogues.
Currently clinical documentation is mainly done by physicians at both steps, a labor intensive process that contributes to physician burnout, defined as a state of emotional, physical, and mental exhaustion caused by prolonged stress in the workplace ~\cite{ortega2023patterns,budd2023burnout}.  
In this paper, we introduce NoteChat, a novel cooperative multi-agent framework leveraging Large Language Models (LLMs) to generate patient-physician conversations conditioned on clinical notes. NoteChat has the potential to help clinical documentation at both steps. 

\begin{table}[H]
\centering
\vspace{-3mm}
\scalebox{0.75}{
\begin{tabular}{l|ccc|cc}
\hline
& \tiny{\textbf{Ours-PMC}} & \tiny{ChatDoctor} & \tiny{DoctorGLM} & \tiny{\textbf{Ours-MTS}} &
\tiny{MTS-Dialog} \\
\hline
\small{\#dial.}     &30k & 112k & 3.4M & 20 & 87\\
\small{\#utt.}       &633k & 224k & 11.2M & 1.25k & 4.79k \\
\small{Chat}          & \small{\cmark} & \small{\xmark} & \small{\xmark} & \small{\cmark} & \small{\cmark}\\
\small{Note}          & \small{\cmark} & \small{\xmark} & \small{\xmark} & \small{\cmark} & \small{\cmark}\\
\small{Syn.}          & \small{AI} & \small{\xmark} & \small{\xmark} & \small{AI} & \small{Human} \\
\small{Lang}          & \small{EN} & \small{EN} & \small{CN} & \small{EN} & \small{EN} \\
\hline
&\multicolumn{5}{c}{\# of utterances in a dialogue} \\
\hline
\small{Avg}        &21.1 & 2 & 3.3 & 62.5 & 55.1 \\
\small{Max}        &61 & 2 & 198 & 112 & 131\\
\small{Min}        &3 & 2 & 2 & 22 & 7\\
\hline
\end{tabular}
}
\vspace{-3mm}
\caption{
Statistics of our NoteChat dataset and related publicly available resources: PMC-based and MTS-based datasets (OursP and OursM, respectively) and muti-round question answering (Chat).
We use "Note" to determine whether we can generate a full clinical note from the data.
We use "Syn" to determine whether the data is generated (by annotators or AI).
}
\vspace{-6mm}
\end{table}

\begin{figure*}
    \centering
\includegraphics[width=0.8\textwidth]{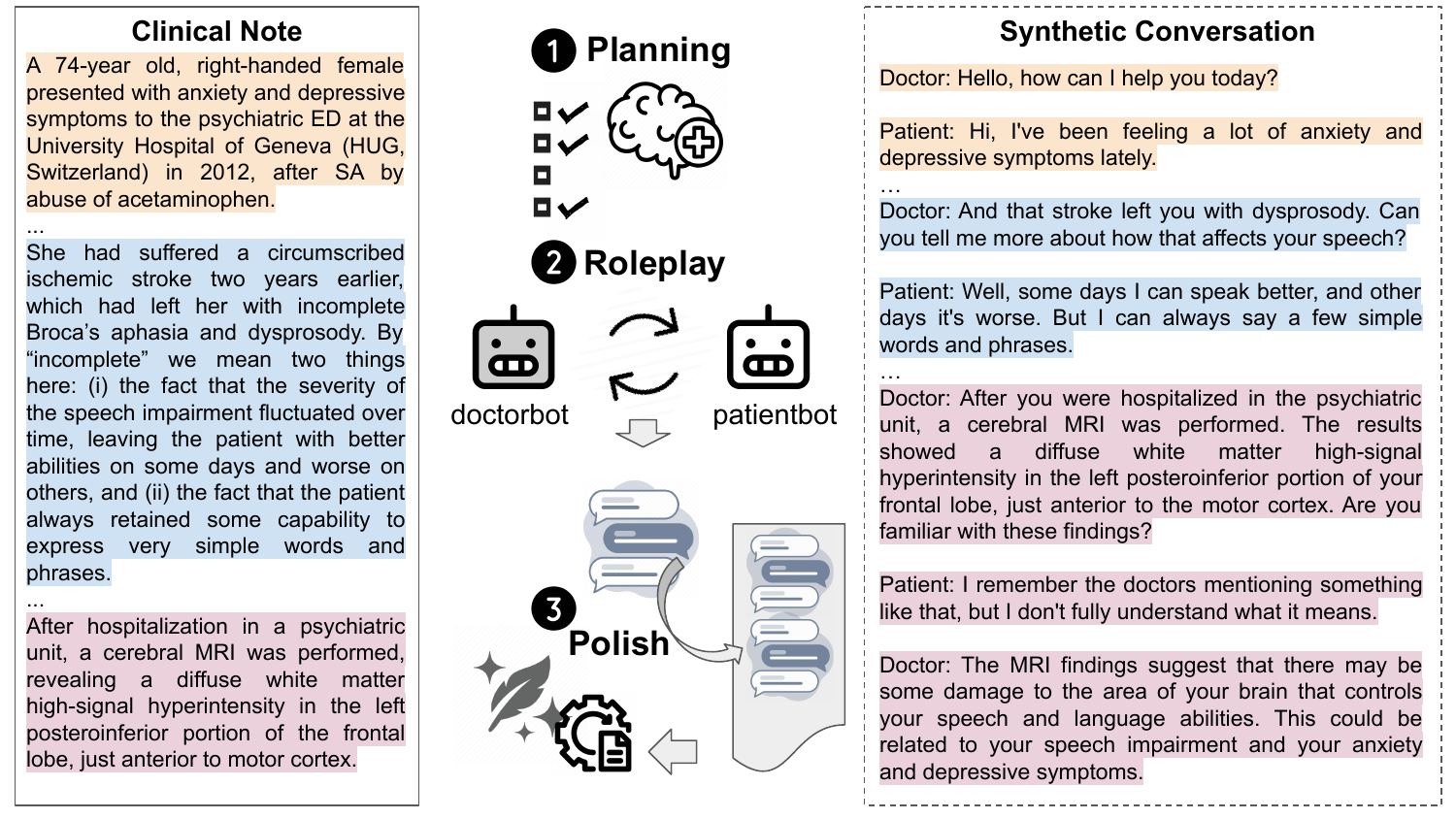}
        \vspace{-4mm}
        \caption{An illustration of NoteChat. 
        \textcolor{Apricot}{Apricot} 
        indicates that our pipeline can generate smooth patient-physician conversations. 
        \textcolor{Cyan}{Blue} shows the characteristics of information seeking, where physicians can actively ask questions to advance the conversation, thanks to \textcircled{2} Roleplay module. In addition, compared with the corresponding note content, the generated utterances are more colloquial, but the key medical concepts are highly overlapped, which reflects NoteChat's control over factuality (mainly from \textcircled{1} Planning module). 
        \textcolor{Lavender}{Lavender} means that NoteChat can generate reasonable explanations for patients, and a lot of information in the chat is reasonable imagination instead of hallucination. 
        The two modules of \textcircled{2} Roleplay and \textcircled{3} Polish can stimulate the imaginative potential of LLMs and reduce unreasonable hallucination through self-examination.} 
        \label{fig:doctor_noteaid}
        \vspace{-4mm}
\end{figure*}

NoteChat leverages LLMs, powerful artificial intelligence (AI) systems extensively trained on a large amount of textual data which represent a significant breakthrough in AI~\cite{brown2020language,longpre2023flan}.
The GPT series by OpenAI~\cite{openai2023gpt4} have demonstrated impressive outcomes and hold significant potential in revolutionizing a broad range of sectors, including marketing, education, and customer service. However, recent work~\cite{mediqa-chat-2023} found ChatGPT does not perform well enough in generating either patient-physician encounter conversation or its corresponding EHR notes. The exploration of open-source LLMs (e.g., LLaMA2) \cite{touvron2023llama, alpaca, vicuna2023} in the medical field remains relatively untapped~\cite{gilson2023does}, despite their immense potential for transforming healthcare communication and decision-making~\cite{abacha2015means}. We suspect that one main reason is the lack of high-quality medical datasets that meet various needs.

Although efforts have been made to create benchmark datasets, the datasets relevant to clinical documentation are small scale~\cite{abacha2023empirical, mediqa-chat-2023, aci-demo}. 
~\citet{yunxiang2023chatdoctor} collected 100k real-world patient-physician conversations from online medical consultation websites as ChatDoctor dataset.
~\citet{xiong2023doctorglm} converted the ChatDoctor data into Chinese and additionally added relevant Chinese dialogue~\cite{zeng2020meddialog} and question-answering. However, none of the aforementioned datasets include dialogue-note pairs. Moreover, 
as indicated in Table 1, the maximum average number of utterances in the existing datasets ~\cite{zeng2020meddialog}
is 3.3, which is a typical representation of online medical consultation websites but markedly less than face-to-face communication between patient and physician encounters~\cite{drew2001conversation}.

The primary challenge of creating benchmark datasets in the clinical domain is HIPAA regulation ~\cite{rindfleisch1997privacy, annas2003hipaa}.  This impediment prevents the use of state-of-the-art LLMs, such as GPTs, on real patient data.  NoteChat circumvents it by generating high-quality synthetic patient-physician conversations conditioned on clinical notes. This synthetic dialogue data can then be used to help train downstream tasks such as clinical note generation conditioned on patient-physician dialogues. Therefore, NoteChat helps both steps of clinical documentation, this is in contrast to the existing models, which mainly focused on clinical dialogue generation only~\cite{yunxiang2023chatdoctor, zeng2020meddialog}.

In this study, we introduce NoteChat, which is built upon a novel cooperative multi-agent framework to generate synthetic patient-physician conversations conditioned on clinical documents (e.g., HIPAA-compliant clinical notes~\footnote{\url{https://github.com/abachaa/MTS-Dialog}} and case reports~\footnote{\url{https://github.com/zhao-zy15/PMC-Patients}}).
NoteChat comprises three modules: Planning, Roleplay, and Polish.
The planning module is responsible for knowledge organization, aiming to decrease hallucination and enhance the consistency of medical logic.
The Roleplay module includes two ChatGPT agents \footnote{We use OpenAI's GPT-3.5 model \texttt{gpt-3.5-turbo-0613}.} take on the roles of physician and patient, respectively. This setup facilitates the generation of interactive dialogues in a looped format.
The Polish module is then utilized to refine these dialogues, ensuring they are more closely aligned with the expectations and preferences of medical professionals, following the feedback and suggestions obtained from physicians and medical students.
Extensive automatic and human evaluations demonstrate the efficacy of our cooperative multi-agent framework and show that NoteChat holds great promise for promoting high-quality synthetic patient-physician conversations.

\noindent{\textbf{In summary, our contributions are as follows:}}
\begin{itemize}
[leftmargin=.2in,topsep=0.1pt]
\setlength\itemsep{0.0em}
\vspace{-0.2em}
    \item We created a novel multiple roleplay LLMs cooperating framework and successfully deployed the framework for the task of generating patient-physician conversations conditioning on clinical notes.
    Although synthetic data generation is an active field in the clinical domain especially to overcome privacy concerns~\cite{pereira2022secure,shafquat2022source,mishra2023synthetic}, to our knowledge, this is the first work to present an instance of multiple LLMs cooperating~\cite{Li2023CAMELCA} to complete a patient-physician conversation conditioned on clinical notes.
    \item We evaluated the quality of the synthetic patient-physician conversations generated by NoteChat with the state-of-the-art OpenAI's ChatGPT and GPT-4 using extensive intrinsic and extrinsic evaluation methods. Through comprehensive human evaluations, we demonstrate that NoteChat holds promise to generate high-quality synthetic patient-physician dialogues.
    \item In this study, we released the first large and high-quality synthetic dialogue data conditioned on 167k case reports that can be used to train both dialogue systems and EHR note-generation systems using dialogues.

    
\end{itemize}

\section{Methods}

\subsection{Data Resource and Preprocessing}

\paragraph{PMC-Patients} is a comprehensive dataset comprising 167K patient case reports and relations extracted from a diverse range of case reports available in the PubMed Central (PMC) repository~\cite{zhao2023pmcpatients}.
PMC-Patient dataset encompasses a vast array of case reports, many of which pertain to rare conditions. To maintain the quality of the generated dialogue in our study, we instruct ChatGPT to exclude exceptionally rare cases. Furthermore, we also instruct ChatGPT to omit case reports related to animal diseases, as they typically bear less relevance to our objective of focusing on human clinical dialogues.

\paragraph{MTS-Dialog} 
is a new collection~\cite{abacha2023empirical, mediqa-chat-2023} of 1.7k short patient-physician conversations and corresponding summaries with section headers and contents following SOAP format~\cite{podder2021soap} to foster advancements in the field of automatic clinical note generation from patient-physician conversations.
This 1.7k short version dataset has a corresponding long version~\cite{aci-demo} of 87 complete dialogues and clinical notes, all of which we use for our evaluation.
However, due to the API's stringent maximum token restriction, incorporating the complete dialogue into a single prompt proved impracticable. 
Consequently, we implemented a strategy that involved segmenting a clinical note into several sections according to the traditional SOAP format~\footnote{SOAP structure details can be found in the Appendix \ref{SOAP-Structure-appendix}.}. 
We used each section header to construct a distinct prompt with the corresponding content in the note, thereby aiding the model in generating individual chats for every section. 
We added a corresponding postprocessing step for MTS-Dialog with the Combine Prompt in Appendix Table~\ref{combine_prompt}, where we concatenated all the small chats from different sections to create a complete dialogue.

\subsection{NoteChat: Generating patient-physician dialogues from notes in the GPT Era}

To ensure that our synthetic datasets closely resemble authentic dialogues, we first use the prompts in Appendix~\ref{baseline_prompt} to guide the roleplay of ChatGPT and GPT4 in generating high-quality data as our baselines.
In this section, we introduce our NoteChat Framework for this task. All our NoteChat experiments in this paper are based on ChatGPT API (gpt-3.5-turbo), but NoteChat can be used in any model that can handle the instructions.





\subsubsection{Main dialogue generation loop} 

\paragraph{Planning module} 

Typically, a physician's diagnostic process adheres to a logical sequence, which may be outlined as follows~\cite{first2013quality, johnson2003critical, tsichlis2021past}: 1) Eliciting symptoms, such as chest pain, 2) Inquiring about the duration of these symptoms, 3) Obtaining medical history, including personal and familial records, 4) Conducting diagnostic tests, 5) Reaching a conclusion and prescribing appropriate medication. 
Thus, an effective dialogue dataset should accurately reflect the logical sequence of real-world interactions between physicians and patients. 
Therefore, before generating dialogues, it is crucial to ensure that the model follows such logic. 
However, we found models often tend to overlook crucial information, create hallucination information, or messily skip content that should logically be in the first half of the dialogue and go to generating first with content that should logically appear later.
This is often caused by the LLMs lacking sufficient medical knowledge~\cite{dave2023chatgpt} or low-level planning abilities~\cite{valmeekam2023planning}.

To circumvent these issues, we first extract clinical domain-specific keywords using CUI (Clinical Uniform Identifier) from MedSpaCy \citep{medspacy} with QuickUMLS \citep{Soldaini2016QuickUMLSAF} and require the LLM to build dialogues around these keywords exclusively, where we design the prompt in Appendix Table~\ref{planning_prompt} with the list of keywords to help the LLM generates the dialogue draft.
With this, we inject external clinical knowledge resources for semantic grounding to reduce hallucination.
The Planning module is responsible only for high-level planning, which pertains to the general distribution of different pieces of information within the dialogue. 
However, the control of each specific utterance at a low level is delegated to the Roleplay module (\ref{roleplay}). 
Therefore, the output of the Planning module is not this draft, but a checklist. Each CUI in the checklist is extracted in sequence from the generated draft.
Then, the Planning module will accompany the entire Roleplay module. That is, every time the Roleplay module completes a new round of dialogue generation, the planning module will count the newly added CUIs in the dialogue and remove them from the checklist. 
Therefore, the Planning module not only assumes the responsibility for the correct correlation of the facts but also helps the entire conversation narrow in a more definite direction until the end.

\paragraph{Roleplay module}
\label{roleplay}
The dialogue draft we generated in the Planning module is not high-quality dialogue data. Previous work~\cite{yunxiang2023chatdoctor} shows that dialogues generated by a single LLM often have issues in language diversity and role homogeneity. 
These are manifestations of the shortcomings of LLMs in handling low-level planning for each utterance in an entire dialogue.
Therefore, in order to generate better quality dialogues, we use the checklist in the Planning module to generate multiple rounds of dialogues using two LLMs to play the roles of patients and physicians, respectively.
This strategy enables us to use distinct prompts based on different requirements of the corresponding role so that the physician's responses appear more professional and the patient's dialogue sounds more normal.
Furthermore, we can control the direction of each dialogue round by modifying the prompts. More specifically, we determine the keywords covered in each round based on the current checklist, allowing two roleplay LLMs to advance the dialogue further and maximize the coverage of the keywords. We then let the Planning module update the checklist.
Subsequently, we let the patient-LLM respond to the physician in as colloquial a manner as possible, ensuring the patient's utterance lay language style. 
All prompts can be found in Table~\ref{roleplay_prompt}.

\paragraph{Polish module}

    \begin{table}
    \centering
    \scalebox{0.75}{
    \begin{tabular}{l|ccc}
    \hline
    & \small{NoteChat} & \small{ChatGPT} & \small{GPT4}\\
    \hline
    \small{total \#dial.}     & 10k & 10k & 10k \\
    \hline
    &\multicolumn{3}{c}{avg \# in a dialogue} \\
    \hline
    \small{utterance}  & 25.4 & 20.5 & 17.4  \\
    \small{word}       &534 & 352 & 390  \\
    \small{medical.}       &59.70 & 44.5 & 51.2  \\
    \hline
    &\multicolumn{3}{c}{avg \# of words in an utterance} \\
    \hline
    \small{physician}          & 30.2 & 25.1 & 33.6 \\
    \small{patient}          & 12.0 & 11.7 & 9.4 \\
    \hline
    &\multicolumn{3}{c}{avg medical term density \%} \\
    \hline
    \small{physician}          & 15.3 & 15.0 & 16.9 \\
    \small{patient}          & 11.2 & 13.4 & 13.0 \\
    \hline
    \end{tabular}
    }
    \caption{Statistics of three synthetic patient-physician dialogue datasets conditioned on PMC-Patient notes~\footnote{For the issue of cost and rate limit, up to now, we only have all 167k ChatGPT synthetic data and only generated 10k synthetic data for GPT4 and NoteChat. For a fair comparison, all experiments and statistics in this paper are based on the same 10k PMC-Patient notes. But we will generate and release all 167k data in the future.}. In the table, we bifurcated the dialogue into two constituent segments: one representing the physician and the other the patient, for which we separately computed their corresponding scores. We computed the average count of words in both the physician and patient utterances across each dialogue in the triad of datasets. Additionally, we derived a metric, indicated as medical term density, which signifies the proportion of the count of Clinical Uniform Identifier (CUI) codes encapsulated within each utterance of physician and patient to the overall count of words.}
    \vspace{-6mm}
    \end{table}

Although the two modules of Planning and Roleplay bring NoteChat more fine-grained control over LLM, restoring patient-physician dialogue from clinical notes requires LLM to balance several challenging requirements, including the planning of key information in the clinical note, reasonable information not occurring in the note but would appear in the dialogues, the language style characteristics of different roles, and the authenticity after combining everything into one complete dialogue. 
In the previous Planning and Roleplay modules, LLMs will promote new dialogues based on historical dialogues. 
Inspired by recent work of rethinking and reranking~\cite{gabriel2021discourse, cobbe2021training, ravaut2022summareranker, jiang2022pairreranker, shinn2023reflexion}, we added the Polish module to give LLM another chance for self-reflection and correction post-Roleplay module.
To do this, we invited human experts who summarized the rules based on the preliminary results of NoteChat to help our synthetic data align with experts' preferences, and they came up with 10 special rules: 
1) Make the conversation as colloquial as possible, 
2) Increase the number of rounds of interaction, 
3) Professional terms and vocabulary should come from the physicians, and patients should be more colloquial,
4) Basic symptoms and medical history should come from the patient, not the physician, 
5) The patients' self-reported signs and symptoms should be around the inputs, 
6) Physician inquiries should be logical,
7) If there are multiple consultation records, you can split a conversation into multiple ones and then link them with transfer words (e.g., a few days later), 
8) Range of rounds of interaction,
9) Must contain the given keywords, 
10) Do not generate duplicate information.
Specifically, we added these requirements to the Polish Prompt in Appendix Table~\ref{polish_prompt} and asked the LLM to polish the existing dialogue accordingly. 
We found that multiple iterations of the Polish step can improve the quality of the final synthetic dialogue~\footnote{After balancing the time, cost, and final performance, we set the number of iterations to 2 in our experiments}.

\section{Automatic Evaluation}
MTS-Dialog provides the human-annotated ground truth conversation data for every clinical note, but the PMC-Patient dataset only has case reports. So, we use intrinsic evaluation for MTS-Dialog synthetic data but extrinsic and human evaluation for PMC-Patient synthetic data.

\subsection{Intrinsic Evaluation}
We measure this task of note-to-conversation from four aspects of the MTS-Dialog dataset.


\noindent\textbf{Similarity} We use ROUGE-F1 scores 
~\citep{lin2004rouge} to measure the similarity of the generated conversation and the references.

\noindent\textbf{Factuality}
We follow recent work~\cite{adams2023meta, ramprasad2023generating} using medical concepts to evaluate factuality and make some improvements. Specifically, we use QuickUMLS \citep{Soldaini2016QuickUMLSAF} to extract medical concepts from model-generated dialogues and ground truth dialogues to get two corresponding concept lists.
Then, we calculate the overlap of medical concept lists between two documents, offering insight into the model's grasp of medical knowledge and terminology.
In Table~\ref{intrinsic}, we report the Concept-P/R/F1 as the Factuality metric.

\noindent\textbf{Extractiveness}
We calculate the ROUGE-F1 of src->hypo (clinical note to model-generated dialogue) as our extractiveness metrics to demonstrate how much information in dialogue is extracted from the clinical note. For AI, a shortcut to improve Factuality is to improve Extractiveness. However, recent work shows increasing the factuality by this way might not be ideal in many scenarios~\cite{ladhak2022faithful,goyal2022news}.

\noindent\textbf{Diversity} We use Self-BLEU (SBLEU)~\cite{zhu2018texygen} to evaluate the diversity of the generated conversation for the patient utterances, physician utterances, and overall.

\subsection{Extrinsic Evaluation}

\begin{figure}
    \centering
    \includegraphics[width=0.5\linewidth]{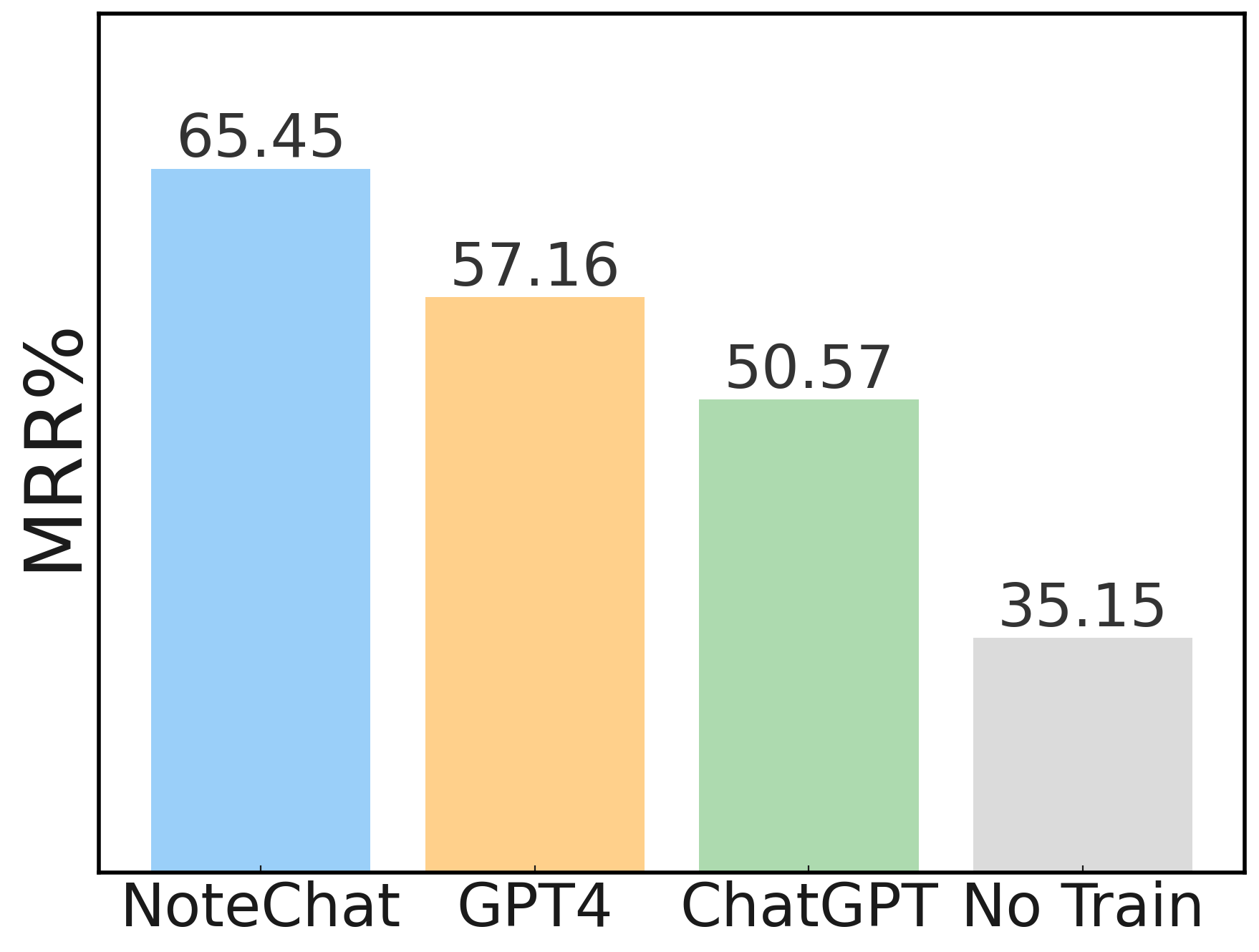}
    \vspace{-3mm}
    \caption{Extrinsic eval results for Medical Chatbot task. LLaMA2-7B is fine-tuned on different PMC-Patient synthetic conversations, and then we use MTS-dialog as the evaluation dataset. NoteChat has the highest score, indicating the most preferred by GPT4.} 
    \label{fig:external}
    \vspace{-3mm}
\end{figure}

\noindent\textbf{Medical Chat Assistant:} We used the PMC-Patient synthetic dialogues generated by ChatGPT, GPT4, and NoteChat to fine-tune the LLaMA2-7B~\footnote{https://huggingface.co/meta-llama/Llama-2-7b-chat-hf}, where we only used physician utterances as the training labels. Then, we evaluated these fine-tuned LLaMA2 chatbots on the ground truth dialogues from MTS-Dialog. For evaluation, recent work shows a higher human evaluation correlation for GPT-4 eval than traditional metrics~\cite{liu2023gpteval, gao2023human, fu2023gptscore, zheng2023judging}, so we also used the GPT4 preference as measurements to evaluate chatbots' response quality.
Specifically, we instruct GPT4 to give preference ranking~\footnote{Prompts can be found in Appendix~\ref{extrinsic_eval_prompt}.} based on the conversation history and the real response. 
We follow~\citet{yao2023improving} to report the Mean Reciprocal Rank (MRR) \cite{radev2002evaluating} of each model's final ranking in Figure~\ref{fig:external}. 
Generally, a higher MRR implies that evaluators have a better alignment with the evaluators' preferences.

\noindent\textbf{Conversation2Note and Note2Conversation:} We also used the NoteChat dataset as data augmentation for two MTS-dialog tasks. We used the same evaluation metrics (ROUGE) following~\citet{mediqa-chat-2023}.

\begin{table}[t!]
\centering
\scalebox{0.6}{
\begin{tabular}{c|ccc}

\hline


\textbf{Similarity} & ROUGE1 & ROUGE2 & ROUGELsum \\
\hline
ChatGPT & 48.56 & 16.74 & 46.36 \\
GPT4 & 53.29 & 20.20 & 50.81 \\
NoteChat & 56.48 & 19.74 & 53.41 \\
\hline


\textbf{Factulity} & Concept-P & Concept-R & Concept-F1 \\
\hline
ChatGPT & 67.54 & 35.75 & 46.23	 \\
GPT4 & 71.46 & 45.69 & 55.17 \\
NoteChat & 48.23 & 51.23 & 49.68 \\
\hline

\textbf{Extractiveness} & src->hypo R1 & src->hypo R2 & src->hypo R-L \\
\hline
ChatGPT & 43.73 & 19.72 & 40.54 \\
GPT4 & 52.70 & 25.70 & 49.63 \\
NoteChat & 37.24 & 20.83 & 36.04 \\
Human & 35.29 & 14.38 & 32.89 \\

\hline


\textbf{Diversity} & all-sbleu $\downarrow$	& physician-sbleu $\downarrow$ & patient-sbleu $\downarrow$ \\
\hline
ChatGPT & 0.017 & 0.006 & 0.017 \\
GPT4 & 0.019 & 0.009 & 0.019 \\
NoteChat & 0.014 & 0.007 & 0.014 \\
\hline

\end{tabular}
}
\vspace{-2mm}
\caption{Intrinsic eval results on MTS-dialog~\tablefootnote{All experiments are done under the zero-shot setting.}.}
\vspace{-4mm}
\label{intrinsic}
\end{table}

\begin{table}[t!]
\centering
\scalebox{0.55}{

\begin{tabular}{@{}lcccc@{}}
\toprule
Model & ROUGE-1 & ROUGE-2 & ROUGE-L & ROUGE-L \\ \midrule
\multicolumn{5}{c}{Note2Conversation} \\
\hline
LLaMA2 (No Train) & 24.60 & 9.26 & 16.19 & 22.92 \\
LLaMA2 (Notechat only) & 36.70 & 22.02 & 29.70 & 35.21 \\
LLaMA2 (MTS only) & 31.09 & 12.80 & 24.30 & 30.05 \\
LLaMA2 (MTS+Notechat) & 42.54 & 19.17 & 38.67 & 38.70 \\
\hline
\multicolumn{5}{c}{Conversation2Note} \\
\hline
LLaMA2 (No Train) & 22.14 & 7.65 & 15.85 & 16.38 \\
LLaMA2 (Notechat only) & 23.82 & 9.08 & 17.37 & 17.48 \\
LLaMA2 (MTS only) & 38.35 & 18.99 & 33.87 & 33.94 \\
LLaMA2 (MTS+Notechat) & 43.84 & 24.34 & 41.05 & 41.06 \\ \bottomrule
\end{tabular}
}
\vspace{-2mm}
\caption{Performance for LLaMA2 fine-tuned on different dataset with Conversation2Note and Note2Conversation extrinsic evaluation tasks.
 }
\vspace{-6mm}
\label{table:external}
\end{table}

\subsection{Automatic Evaluation Results}


The \textbf{intrinsic evaluation} results, as illustrated in Table~\ref{intrinsic}, show that the overall similarity of the conversations generated by NoteChat and Human (MTS-dialog ground truth) is higher than that of GPT4 and ChatGPT baselines.
GPT4 outperformed NoteChat and ChatGPT in both factuality and extractiveness metrics. NoteChat outperformed ChatGPT in factuality but had a lower and closer to human extractiveness score. In Section~\ref{heuristic_eval}, we will discuss the impact of the different factuality and extractiveness scores of the three methods on human expert preferences on our task.
Finally, we found that the diversity of NoteChat, especially for patient utterances, is significantly better than the baselines.
The \textbf{extrinsic evaluation Medical Chat Assistant} results are illustrated in Figure~\ref{fig:external}.
In this experiment, LLaMA2-7B is first fine-tuned on different PMC-Patient synthetic conversations. Then we use MTS-dialog as the evaluation dataset. 
NoteChat-based LLaMA2 has the highest score, indicating the most preferred by GPT4 when generating real physician utterances.
It is worth noting that this evaluation is also a kind of transfer learning because the model is only trained on different versions of PMC-Patient synthetic dialogue (NoteChat, ChatGPT, GPT4) and then tested its zero-shot performance on human-labeled dialogue in MTS-dialog.
The \textbf{extrinsic evaluation Conversation2Note and Note2Conversation} results are illustrated in Table~\ref{table:external}.
We found that training on NoteChat-only can observe significant improvements in MTS-dialogue test results. The best results can be obtained if NoteChat is used as data augmentation of the original MTS-dialogue training data.
Therefore, the results of this extrinsic evaluation show that the models trained on the NoteChat dataset are generalizable to the real human-annotated dataset.

\section{Human Evaluation}

To assess the quality of synthetic conversations generated by different methods (ChatGPT, GPT-4, NoteChat), we conducted a human evaluation using crowd-sourcing and domain experts. 

\subsection{Human Evaluation Settings}

The goal of \textbf{expert evaluation} is to have human domain experts evaluate whether these machine-generated conversations are comparable to real patient-physician encounter conversations from a professional perspective (e.g. medical commonsense, knowledge, logic).
To do so, we recruited 5 medical practitioners\footnote{Four licensed physicians and one medical student with hospital internship experience. These experts were not involved in the research, only the human evaluation.}, and their tasks are to read clinical notes and provide qualitative feedback on whether the machine-generated dialogues can be defined as high-quality patient-physician interactions in terms of factual accuracy and logical coherence; if not, how should they be improved? 

The goal of \textbf{crowd evaluation} is to allow the general public to provide ratings for different synthetic conversations based on their lived experience. Since the crowds do not have professional medical knowledge, participants will first read the clinical notes and medical expert annotated conversations as references for high-quality data and then rank different machine-generated conversations for quantitative measurement of their preference.
We recruited 10 human evaluators to participate in our crowd evaluation. \footnote{All the evaluators have bachelor's degrees but do not have any medical education background.}

\subsection{Human Evaluation Measurements}
We mainly use human preference as measurements to evaluate synthetic conversation quality.
Specifically, the participants are provided with the following instructions ``\emph{The following three conversations are generated by AI based on this clinical note. Please rank them according to the quality you think, from high to low.}''. 
We collect the preference ranking from experts, crowds, and GPT4.
We report the Mean Reciprocal Rank (MRR) of each model's final ranking in Figure~\ref{fig:human_eval}.

\subsection{Human Evaluation Outcome}
\label{patient-result}

\begin{figure}[H]
\vspace{-5mm}
    \centering
    \includegraphics[width=0.7\linewidth]{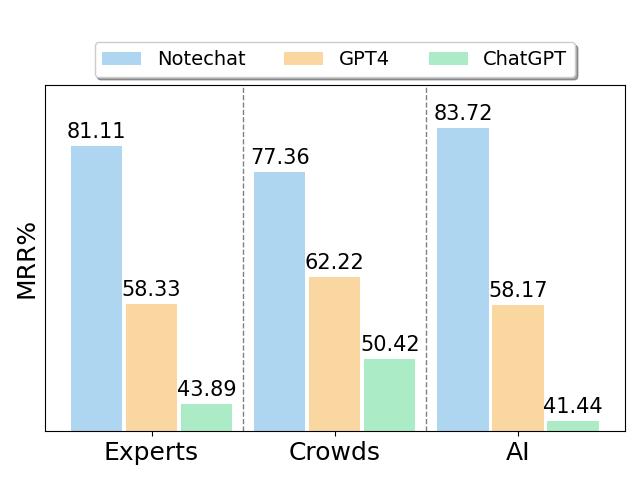}
    \vspace{-4mm}
        \caption{Human\&AI preference for 50 samples.} 
        \label{fig:human_eval}
        \vspace{-4mm}
\end{figure}

All the preference feedback from experts, crowds, and AI are shown in Figure~\ref{fig:human_eval}.
In the most crucial results concerning expert preferences, NoteChat's MRR score significantly outperforms that of GPT4, indicating that from an expert's perspective, the quality of dialogue data from NoteChat is higher. 
In terms of preferences among the crowds and AI, NoteChat also clearly surpasses GPT4, demonstrating consistency with expert preferences. 
Finally, in all three human evaluations, both NoteChat and GPT4 perform better than ChatGPT.

\subsection{Heuristic Evaluation with Experts}
\label{heuristic_eval}

We interviewed 5 medical practitioners: 

Q1) \textbf{What are the shortcomings of AI synthetic conversation compared with real-world patient-physician encounter conversation?}
Experts think that synthetic conversations cover too much information from the clinical note compared to real-world conversations, because some factual information is not provided to note through conversation (such as lab test results). For example, in Table~\ref{tab:example} Example 1, the detailed dosage information will be not in the conversation.
In Example 2, the patient acts too professionally. In the answer, a lot of medical knowledge that physicians will know is described by the patient.

Q2) \textbf{What is the difference between ChatGPT, GPT4, and NoteChat synthetic conversations?}
All medical practitioners believe that GPT4 and NoteChat lead ChatGPT in terms of factuality. Since our NoteChat is based upon ChatGPT, this human observation shows that our modules successfully inject medical concept knowledge to improve the factuality level from ChatGPT to the level of GPT4.
So, as shown in Figure~\ref{fig:human_eval}, ChatGPT is ranked last in all cases.

Regarding the comparison between NoteChat and GPT4, medical practitioners actually believe that the data quality of NoteChat-synthetic conversations is generally better than the GPT4 synthetic dataset, which aligns with their expert preference in Figure~\ref{fig:human_eval}. 
We further conducted a heuristic evaluation to explore the reason here as well as the deficiency of NoteChat and GPT4 synthetic conversations and potential improvement. 
We further conducted a heuristic evaluation to explore the reason here as well as the deficiency of NoteChat and GPT4 synthetic conversations and potential improvement. First of all, GPT4 prefers to copy the information directly in the note to meet the requirements of factuality, but this will make the conversation unreal. In Table~\ref{tab:example} Example 2, the information is highly summarized and put together on the note, but it is unnatural for the same content to appear directly in the dialogue. Compared with the utterance generated by GPT4, a better way is to use multiple conversation rounds to obtain information one by one. This is a problem common to all AIs in this paper, but GPT4's problem is most obvious. Second, in reality, physicians are expected to not only answer questions but also advance the discussion by asking professional questions. We observe that the physician in NoteChat is more likely to advance the conversation compared to the physician in GPT4 due to our Roleplay module.

To better control language models, it's important to specify which information is spoken by the physician and which by the patient. In the Table~\ref{tab:example} Example 3, GPT-4 let the patient speculate about their symptoms and dismiss physical activities as a cause. Using a specific prompt, the NoteChat Roleplay module was adjusted to ensure both the physician and patient roles are accurately portrayed and cooperate logically.
Finally, The dialogue should start like a real conversation, with the patient sharing symptoms and medical history. Usually, doctors don't know a patient's history, so patients need to express or be asked about their symptoms and history. This approach sets the direction for tests and treatment plans. In GPT-4 generated dialogues, this format should be followed, but often, the physician character incorrectly presents this information first, which is not typical in real clinical settings (Example 5 of Table~\ref{tab:example}).

\begin{table}[H]
\vspace{-3mm}
\scalebox{0.8}{
    \begin{scriptsize}
    \setlength{\tabcolsep}{2pt}
    \renewcommand{\arraystretch}{1}
    \centering
    \begin{tabularx}{0.58\textwidth}{lXXX}
    \toprule
    
    \multirow{2}{*}{1} 
    & 
    \textbf{Some information in the note does not come from the conversation, so it is not necessary to mention it in the dialogue, such as dosage information.} \\
    &
    \cellcolor{gray!10} \emph{Her medications were tab. vildagliptin/metformin (50/1000) (mg) BID, tab. amlodipine/valsartan (5/160) (mg) once daily, and tab. levothyroxine 75 mcg once daily ...}\\
      & (\textit{\textcolor{red}{GPT4}}) \textbf{Patient}: I take vildagliptin/metformin 50/1000 mg twice a day, amlodipine/valsartan 5/160 mg once daily, and levothyroxine 75 mcg once daily. \\
      & (\textit{\textcolor{blue}{NoteChat}}) \textbf{Patient}: I have arterial hypertension, diabetes, and hypothyroidism. I take vildagliptin/metformin, amlodipine/valsartan, and levothyroxine. 
 \\
    \midrule
    \multirow{2}{*}{2} 
    &
    \textbf{Patients sometimes act too professionally.} \\
    & 
    \cellcolor{gray!10} \emph{Two months postoperatively, the patient came to our institution and underwent MRI for reevaluation of the cervical spine in three different positions: neutral, flexion, and extension. While neutral MRI did not show any significant stenosis, flexion and extension MRIs ...} \\
    & (\textit{\textcolor{red}{GPT4}})
    \textbf{Patient}: The MRI was done in three different positions: neutral, flexion, and extension. The neutral position didn't show any significant stenosis, but the flexion and extension MRIs showed cervical compression due to spondylosis... \\
    & (\textit{\textcolor{blue}{NoteChat}}) \textbf{physician:} Thank you. Let me review your medical history records. (physician reviews the medical history records) Based on your medical history records, it seems that the previous surgery did not completely resolve your symptoms. We will need to perform a reevaluation of your cervical spine to determine the cause of your symptoms.. 
    \textbf{Patient:} Okay, I understand. 
 \\ 
    \midrule
    \multirow{2}{*}{3} 
    &
    \textbf{Highly generalized sentences in notes tend to contain a lot of important information, but in dialogues, they are usually collected by multiple rounds.} \\
    & 
    \cellcolor{gray!10} \emph{Her diagnosis at admission included a gallstone, acute attack of chronic cholecystitis, biliary pancreatitis, septic shock, and multiple organ dysfunction syndrome ...} \\
    & (\textit{\textcolor{red}{GPT4}}) \textbf{physician}: So, I see from your chart that you were admitted with a gallstone, acute attack of chronic cholecystitis, biliary pancreatitis, septic shock, and multiple organ dysfunction syndrome. Can you tell me more about your symptoms? \textbf{Patient}: Well, I had severe abdominal pain and vomiting. \\
    & (\textit{\textcolor{blue}{NoteChat}}) 
    \textbf{physician:} Great, I appreciate your cooperation. Now, could you please provide me with your medical history records?
    \textbf{Patient:} Sure, here they are.
    \textbf{physician:} Thank you. Based on your medical history records, it seems that you were admitted with a diagnosis of a gallstone, an acute attack of chronic cholecystitis, biliary pancreatitis...
    \\
    \midrule
    \multirow{2}{*}{4} 
    &
    \textbf{The logic of the different roles in the dialogue. It is necessary to distinguish which information will be advanced by the physician and which information will be provided by the patient.}
    \\
    &
    \cellcolor{gray!10} \emph{A 66-year-old woman presented at the emergency department of our hospital complaining of sudden onset dizziness and fatigue over the past thirty minutes. Before her arrival, she was at home relaxing and not engaged in any physical activity ...} \\
    & (\textit{\textcolor{red}{GPT4}}) \textbf{Patient}: Hi, physician. About thirty minutes ago, I suddenly started feeling dizzy and fatigued while I was at home relaxing. \textcolor{orange}{I wasn't even engaged in any physical activity.} \textbf{physician}: I see. Can you please tell me about your past medical history?\\
    & (\textit{\textcolor{blue}{NoteChat}}) \textbf{physician}: I see. Can you tell me more about your symptoms? \textcolor{orange}{Were you relaxing or engaged in any physical activity before the symptoms occurred?} \textbf{Patient}: I was just at home relaxing before the symptoms started.
 \\
    \midrule
    \multirow{2}{*}{5} 
    & 
    \textbf{NoteChat handles the beginning of the conversation more naturally. So experts prefer NoteChat when other aspects of quality are similar.} \\
    & 
    \cellcolor{gray!10} \emph{A 71-year-old man with a history of atrial fibrillation, acute appendicitis, and early esophageal cancer treated with endoscopic submucosal dissection underwent distal pancreatectomy ...} \\
    & (\textit{\textcolor{red}{GPT4}}) \textbf{physician:} Hello, I see from your history that you have atrial fibrillation, acute appendicitis, and early esophageal cancer. You were treated with endoscopic submucosal dissection for esophageal cancer, correct?
    \textbf{Patient}: Yes, that's right. \\
    & (\textit{\textcolor{blue}{NoteChat}}) \textbf{Patient:} physician, hello. I have an irregular posterior wall and a submucosal tumor in the anterior wall of my gastric antrum.
    \textbf{physician:} Can you give me your medical records?
    \textbf{Patient:} Here you go.
 \\
    \bottomrule
    \end{tabularx}
    \end{scriptsize}
    }
\vspace{-2mm}
\caption{Expert evaluation case study~\tablefootnote{Due to the obvious gap in factuality of ChatGPT, our cases focus on the difference between NoteChat and GPT4.}.}
\label{tab:example}
\vspace{-8mm}
\end{table}

\section{Related Work}

\textbf{Clinical note and conversations generation:}
A task closely related to our work, but with an inverse direction, is the automatic generation of clinical notes from patient-physician conversations~\cite{krishna2020generating, song-summarizing,  yim2021towards, Su2022ExtractAA, yao2023improving}.
Recently, the MEDIQA-Chat 2023 \footnote{https://sites.google.com/view/mediqa2023} introduced tasks in both directions (Dialogue2Note Summarization and Note2Dialogue Generation).
However, their dataset is either private or limited to less than 2k examples. 
One of the main themes of recent data-centric AI is the synthetic data to overcome privacy concerns~\cite{pereira2022secure,shafquat2022source,mishra2023synthetic}.
To the best of our knowledge, we are the first to introduce a large-scale publicly available patient-physician conversation dataset in English, each accompanied by corresponding medical documents, with an average number of utterances exceeding 20 rounds.
In addition, our extrinsic eval shows that the NoteChat can be used as auxiliary data for Conversation2Note or Note2Conversation tasks and can also be used as a synthetic medical dialogue dataset alone to engage patients directly and help clinical documentation~\cite{huatuogpt-2023, li2023huatuo26m, wang2023huatuo, MedicalGPT-zh, xiong2023doctorglm, zeng2020meddialog}.

\noindent\textbf{Multiple LLMs cooperation:}
Our work builds upon the recent advances in deploying two LLMs as cooperative agents \citep{Panait2005CooperativeML} for multi-round conversation generation. 
In particular, NoteChat is inspired by CAMEL \citep{Li2023CAMELCA}, which assigns roles to two LLMs (e.g. student and teacher) in order to facilitate conversation between the two agents for a particular task (e.g. teaching). Similar to CAMEL's findings, we found that roleplay by itself may hallucinate 
or generate fake replies that repeat most of the previous utterances. To solve this issue, we proposed a novel Planning module to ground agents to certain keywords. 
\citet{Cho2023WhenCM} also addresses the challenges of using LLM to craft a dialogue dataset with specified personas. They emphasize the importance of grounding and context in conversation generation. Similarly, NoteChat relies on structured clinical notes segmented using the SOAP format to provide context for our dialogue synthesis to diagnose a patient. However, their work is limited to generating open-domain dialogue, while we focus on task-oriented dialogue.

\section{Conclusion}
\label{sec:conclusion}
In this study, we present \textbf{\emph{NoteChat}}, a cooperative multi-agent framework leveraging LLMs for generating synthetic patient-physician conversations conditioned on clinical notes. NoteChat consists of Planning, Roleplay, and Polish modules. Extensive evaluations demonstrate that NoteChat facilitates high-quality synthetic patient-physician conversations, underscoring the untapped potential of LLMs in healthcare and offering promising avenues for the intersection of AI and healthcare.

\section{Limitations and Ethical Considerations}
\label{sec:ethics}
This study offers valuable insights, but with a few limitations, we would like to note.

Due to cost and time constraints, we could not try out many possibilities and alternatives in this paper. First of all, the current amount of data for human evaluation is not particularly sufficient. We are conducting more human evaluations. Secondly, due to cost issues, we currently do not use GPT-4 extensively to try the NoteChat pipeline. When OpenAI updates the Stateful API~\footnote{https://www.reuters.com/technology/openai-plans-major-updates-lure-developers-with-lower-costs-sources-2023-10-11/}, we will use this version to generate NoteChat-GPT4. Third, we extracted relevant UMLS-CUI codes for our Planning module, aiming to guide subsequent conversations around these critical terms. Such a checklist can help our pipeline improve factuality~\cite{asai2023retrieval, Huang2023ASO}, and can be very flexibly combined with other tools to meet different purposes, like information retrieval~\cite{khattab2022demonstrate}, entity\&relation extraction~\cite{cai2023paniniqa}, medical jargon extraction~\cite{kwon2022medjex}, causal inference~\cite{yuan2023causality}, evidence and reasoning path retrieval~\cite{asai2019learning, asai2021evidentiality}, and many other knowledge injection ideas~\cite{fei2021enriching, yao2021improving}.

Consider Privacy Implications, LLMs can present privacy concerns in using clinical notes to generate patient-physician conversation, potentially violating HIPAA regulations. However, in this study, all experiments were sourced from publicly available real patient data collected from research articles with at least CC BY-NC-SA license. We also present an approach for generating synthetic conversations from case reports in the PubMed Central repository.

Consider Biases, LLMs trained on vast amounts of text data may inadvertently capture and reproduce biases present in the data. For example, they may prefer certain questions related to Metformin or link particular health conditions to specific populations. Thus the physician bot trained from our synthetic data may perpetuate incorrect information or provide inaccurate answers. Moreover, the case reports used to generate synthetic conversations usually focus on unusual observations and rare conditions. Thus the physician bot may hallucinate or overtreat patients with common diseases.

Considering Broader Impacts, we have performed a preliminary study to generate synthetic conversation from case reports within research articles indexed from January 2002 to July 2022 by PubMed Central. The credibility of these case reports is ensured as they are peer-reviewed and published in academic journals. Moreover, the type of disease is diverse as they are sourced from various hospital departments and are not limited to intensive care units (such as MIMIC). Thus, models trained using our synthetic data may benefit from these characteristics.


\bibliography{emnlp23}

\begin{thebibliography}{69}
\expandafter\ifx\csname natexlab\endcsname\relax\def\natexlab#1{#1}\fi

\bibitem[{Abacha et~al.(2023)Abacha, Yim, Fan, and Lin}]{abacha2023empirical}
Asma~Ben Abacha, Wen-wai Yim, Yadan Fan, and Thomas Lin. 2023.
\newblock An empirical study of clinical note generation from doctor-patient
  encounters.
\newblock In \emph{Proceedings of the 17th Conference of the European Chapter
  of the Association for Computational Linguistics}, pages 2283--2294.

\bibitem[{Abacha and Zweigenbaum(2015)}]{abacha2015means}
Asma~Ben Abacha and Pierre Zweigenbaum. 2015.
\newblock Means: A medical question-answering system combining nlp techniques
  and semantic web technologies.
\newblock \emph{Information processing \& management}, 51(5):570--594.

\bibitem[{Adams et~al.(2023)Adams, Zucker, and Elhadad}]{adams2023meta}
Griffin Adams, Jason Zucker, and No{\'e}mie Elhadad. 2023.
\newblock A meta-evaluation of faithfulness metrics for long-form
  hospital-course summarization.
\newblock \emph{arXiv preprint arXiv:2303.03948}.

\bibitem[{Annas(2003)}]{annas2003hipaa}
George~J Annas. 2003.
\newblock Hipaa regulations: a new era of medical-record privacy?
\newblock \emph{New England Journal of Medicine}, 348:1486.

\bibitem[{Asai et~al.(2021)Asai, Gardner, and
  Hajishirzi}]{asai2021evidentiality}
Akari Asai, Matt Gardner, and Hannaneh Hajishirzi. 2021.
\newblock Evidentiality-guided generation for knowledge-intensive nlp tasks.
\newblock \emph{arXiv preprint arXiv:2112.08688}.

\bibitem[{Asai et~al.(2019)Asai, Hashimoto, Hajishirzi, Socher, and
  Xiong}]{asai2019learning}
Akari Asai, Kazuma Hashimoto, Hannaneh Hajishirzi, Richard Socher, and Caiming
  Xiong. 2019.
\newblock Learning to retrieve reasoning paths over wikipedia graph for
  question answering.
\newblock \emph{arXiv preprint arXiv:1911.10470}.

\bibitem[{Asai et~al.(2023)Asai, Min, Zhong, and Chen}]{asai2023retrieval}
Akari Asai, Sewon Min, Zexuan Zhong, and Danqi Chen. 2023.
\newblock Retrieval-based language models and applications.
\newblock In \emph{Proceedings of the 61st Annual Meeting of the Association
  for Computational Linguistics (Volume 6: Tutorial Abstracts)}, pages 41--46.

\bibitem[{{Ben Abacha} et~al.(2023){Ben Abacha}, Yim, Adams, Snider, and
  Yetisgen}]{mediqa-chat-2023}
Asma {Ben Abacha}, Wen{-}wai Yim, Griffin Adams, Neal Snider, and Meliha
  Yetisgen. 2023.
\newblock Overview of the mediqa-chat 2023 shared tasks on the summarization
  and generation of doctor-patient conversations.
\newblock In \emph{ACL-ClinicalNLP 2023}.

\bibitem[{Brown et~al.(2020)Brown, Mann, Ryder, Subbiah, Kaplan, Dhariwal,
  Neelakantan, Shyam, Sastry, Askell, Agarwal, Herbert-Voss, Krueger, Henighan,
  Child, Ramesh, Ziegler, Wu, Winter, Hesse, Chen, Sigler, Litwin, Gray, Chess,
  Clark, Berner, McCandlish, Radford, Sutskever, and
  Amodei}]{brown2020language}
Tom~B. Brown, Benjamin Mann, Nick Ryder, Melanie Subbiah, Jared Kaplan,
  Prafulla Dhariwal, Arvind Neelakantan, Pranav Shyam, Girish Sastry, Amanda
  Askell, Sandhini Agarwal, Ariel Herbert-Voss, Gretchen Krueger, Tom Henighan,
  Rewon Child, Aditya Ramesh, Daniel~M. Ziegler, Jeffrey Wu, Clemens Winter,
  Christopher Hesse, Mark Chen, Eric Sigler, Mateusz Litwin, Scott Gray,
  Benjamin Chess, Jack Clark, Christopher Berner, Sam McCandlish, Alec Radford,
  Ilya Sutskever, and Dario Amodei. 2020.
\newblock \href
  {https://proceedings.neurips.cc/paper/2020/file/1457c0d6bfcb4967418bfb8ac142f64a-Paper.pdf}
  {Language models are few-shot learners}.
\newblock In \emph{Advances in Neural Information Processing Systems},
  volume~33, pages 1877--1901.

\bibitem[{Budd(2023)}]{budd2023burnout}
Jeffrey Budd. 2023.
\newblock Burnout related to electronic health record use in primary care.
\newblock \emph{Journal of Primary Care \& Community Health},
  14:21501319231166921.

\bibitem[{Cai et~al.(2023)Cai, Yao, Liu, Wang, Reilly, Zhou, Li, Cao, Kapoor,
  Bajracharya et~al.}]{cai2023paniniqa}
Pengshan Cai, Zonghai Yao, Fei Liu, Dakuo Wang, Meghan Reilly, Huixue Zhou,
  Lingxi Li, Yi~Cao, Alok Kapoor, Adarsha Bajracharya, et~al. 2023.
\newblock Paniniqa: Enhancing patient education through interactive question
  answering.
\newblock \emph{arXiv preprint arXiv:2308.03253}.

\bibitem[{Chiang et~al.(2023)Chiang, Li, Lin, Sheng, Wu, Zhang, Zheng, Zhuang,
  Zhuang, Gonzalez, Stoica, and Xing}]{vicuna2023}
Wei-Lin Chiang, Zhuohan Li, Zi~Lin, Ying Sheng, Zhanghao Wu, Hao Zhang, Lianmin
  Zheng, Siyuan Zhuang, Yonghao Zhuang, Joseph~E. Gonzalez, Ion Stoica, and
  Eric~P. Xing. 2023.
\newblock \href {https://vicuna.lmsys.org} {Vicuna: An open-source chatbot
  impressing gpt-4 with 90\%* chatgpt quality}.

\bibitem[{Cho et~al.(2023)Cho, Lee, Bae, Kim, Park, Kim, Hahn, and
  Kim}]{Cho2023WhenCM}
Won~Ik Cho, Yoon~Kyung Lee, Seoyeon Bae, Ji-Hwan Kim, Sangah~Nancy Park,
  Moosung Kim, Sowon Hahn, and Nam~Soo Kim. 2023.
\newblock When crowd meets persona: Creating a large-scale open-domain persona
  dialogue corpus.
\newblock \emph{ArXiv}, abs/2304.00350.

\bibitem[{Cobbe et~al.(2021)Cobbe, Kosaraju, Bavarian, Chen, Jun, Kaiser,
  Plappert, Tworek, Hilton, Nakano et~al.}]{cobbe2021training}
Karl Cobbe, Vineet Kosaraju, Mohammad Bavarian, Mark Chen, Heewoo Jun, Lukasz
  Kaiser, Matthias Plappert, Jerry Tworek, Jacob Hilton, Reiichiro Nakano,
  et~al. 2021.
\newblock \href {https://arxiv.org/abs/2110.14168} {Training verifiers to solve
  math word problems}.
\newblock \emph{ArXiv preprint}, abs/2110.14168.

\bibitem[{Dave et~al.(2023)Dave, Athaluri, and Singh}]{dave2023chatgpt}
Tirth Dave, Sai~Anirudh Athaluri, and Satyam Singh. 2023.
\newblock Chatgpt in medicine: an overview of its applications, advantages,
  limitations, future prospects, and ethical considerations.
\newblock \emph{Frontiers in Artificial Intelligence}, 6:1169595.

\bibitem[{Drew et~al.(2001)Drew, Chatwin, and Collins}]{drew2001conversation}
Paul Drew, John Chatwin, and Sarah Collins. 2001.
\newblock Conversation analysis: a method for research into interactions
  between patients and health-care professionals.
\newblock \emph{Health Expectations}, 4(1):58--70.

\bibitem[{Eyre et~al.(2021)Eyre, Chapman, Peterson, Shi, Alba, Jones, Box,
  DuVall, and Patterson}]{medspacy}
H.~Eyre, A.~B. Chapman, K.~S. Peterson, J.~Shi, P.~R. Alba, M.~M. Jones, T.~L.
  Box, S.~L. DuVall, and O.~V. Patterson. 2021.
\newblock {{L}aunching into clinical space with medspa{C}y: a new clinical text
  processing toolkit in {P}ython}.
\newblock \emph{AMIA Annu Symp Proc}, 2021:438--447.

\bibitem[{Fei et~al.(2021)Fei, Ren, Zhang, Ji, and Liang}]{fei2021enriching}
Hao Fei, Yafeng Ren, Yue Zhang, Donghong Ji, and Xiaohui Liang. 2021.
\newblock Enriching contextualized language model from knowledge graph for
  biomedical information extraction.
\newblock \emph{Briefings in bioinformatics}, 22(3):bbaa110.

\bibitem[{First et~al.(2013)First, Chaudhry, and Melnick}]{first2013quality}
Lewis~R First, Humayun~J Chaudhry, and Donald~E Melnick. 2013.
\newblock Quality, cost, and value of clinical skills assessment.
\newblock \emph{New England Journal of Medicine}, 368(10):963--964.

\bibitem[{Fu et~al.(2023)Fu, Ng, Jiang, and Liu}]{fu2023gptscore}
Jinlan Fu, See-Kiong Ng, Zhengbao Jiang, and Pengfei Liu. 2023.
\newblock Gptscore: Evaluate as you desire.
\newblock \emph{arXiv preprint arXiv:2302.04166}.

\bibitem[{Gabriel et~al.(2021)Gabriel, Bosselut, Da, Holtzman, Buys, Lo,
  Celikyilmaz, and Choi}]{gabriel2021discourse}
Saadia Gabriel, Antoine Bosselut, Jeff Da, Ari Holtzman, Jan Buys, Kyle Lo,
  Asli Celikyilmaz, and Yejin Choi. 2021.
\newblock \href {https://doi.org/10.18653/v1/2021.eacl-main.34} {Discourse
  understanding and factual consistency in abstractive summarization}.
\newblock In \emph{Proceedings of the 16th Conference of the European Chapter
  of the Association for Computational Linguistics: Main Volume}, pages
  435--447, Online. Association for Computational Linguistics.

\bibitem[{Gao et~al.(2023)Gao, Ruan, Sun, Yin, Yang, and Wan}]{gao2023human}
Mingqi Gao, Jie Ruan, Renliang Sun, Xunjian Yin, Shiping Yang, and Xiaojun Wan.
  2023.
\newblock Human-like summarization evaluation with chatgpt.
\newblock \emph{arXiv preprint arXiv:2304.02554}.

\bibitem[{Gilson et~al.(2023)Gilson, Safranek, Huang, Socrates, Chi, Taylor,
  Chartash et~al.}]{gilson2023does}
Aidan Gilson, Conrad~W Safranek, Thomas Huang, Vimig Socrates, Ling Chi,
  Richard~Andrew Taylor, David Chartash, et~al. 2023.
\newblock How does chatgpt perform on the united states medical licensing
  examination? the implications of large language models for medical education
  and knowledge assessment.
\newblock \emph{JMIR Medical Education}, 9(1):e45312.

\bibitem[{Goyal et~al.(2022)Goyal, Li, and Durrett}]{goyal2022news}
Tanya Goyal, Junyi~Jessy Li, and Greg Durrett. 2022.
\newblock News summarization and evaluation in the era of gpt-3.
\newblock \emph{arXiv preprint arXiv:2209.12356}.

\bibitem[{Huang et~al.(2023)Huang, Yu, Ma, Zhong, Feng, Wang, Chen, Peng, Feng,
  Qin, and Liu}]{Huang2023ASO}
Lei Huang, Weijiang Yu, Weitao Ma, Weihong Zhong, Zhangyin Feng, Haotian Wang,
  Qianglong Chen, Weihua Peng, Xiaocheng Feng, Bing Qin, and Ting Liu. 2023.
\newblock \href {https://api.semanticscholar.org/CorpusID:265067168} {A survey
  on hallucination in large language models: Principles, taxonomy, challenges,
  and open questions}.

\bibitem[{Jiang et~al.(2022)Jiang, Lin, and Ren}]{jiang2022pairreranker}
Dongfu Jiang, Bill~Yuchen Lin, and Xiang Ren. 2022.
\newblock Pairreranker: Pairwise reranking for natural language generation.
\newblock \emph{arXiv preprint arXiv:2212.10555}.

\bibitem[{Johnson(2003)}]{johnson2003critical}
Hillary Johnson. 2003.
\newblock A critical review of standardized patient examinations as part of the
  usmle.
\newblock \emph{AMA Journal of Ethics}, 5(12):426--429.

\bibitem[{Khattab et~al.(2022)Khattab, Santhanam, Li, Hall, Liang, Potts, and
  Zaharia}]{khattab2022demonstrate}
Omar Khattab, Keshav Santhanam, Xiang~Lisa Li, David Hall, Percy Liang,
  Christopher Potts, and Matei Zaharia. 2022.
\newblock Demonstrate-search-predict: Composing retrieval and language models
  for knowledge-intensive nlp.
\newblock \emph{arXiv preprint arXiv:2212.14024}.

\bibitem[{Krishna et~al.(2020)Krishna, Khosla, Bigham, and
  Lipton}]{krishna2020generating}
Kundan Krishna, Sopan Khosla, Jeffrey~P Bigham, and Zachary~C Lipton. 2020.
\newblock Generating soap notes from doctor-patient conversations using modular
  summarization techniques.
\newblock \emph{arXiv preprint arXiv:2005.01795}.

\bibitem[{Kwon et~al.(2022)Kwon, Yao, Jordan, Levy, Corner, and
  Yu}]{kwon2022medjex}
Sunjae Kwon, Zonghai Yao, Harmon~S Jordan, David~A Levy, Brian Corner, and Hong
  Yu. 2022.
\newblock Medjex: A medical jargon extraction model with wiki's hyperlink span
  and contextualized masked language model score.
\newblock \emph{arXiv preprint arXiv:2210.05875}.

\bibitem[{Ladhak et~al.(2022)Ladhak, Durmus, He, Cardie, and
  McKeown}]{ladhak2022faithful}
Faisal Ladhak, Esin Durmus, He~He, Claire Cardie, and Kathleen McKeown. 2022.
\newblock \href {https://doi.org/10.18653/v1/2022.acl-long.100} {Faithful or
  extractive? on mitigating the faithfulness-abstractiveness trade-off in
  abstractive summarization}.
\newblock In \emph{Proceedings of the 60th Annual Meeting of the Association
  for Computational Linguistics (Volume 1: Long Papers)}, pages 1410--1421,
  Dublin, Ireland. Association for Computational Linguistics.

\bibitem[{Li et~al.(2023{\natexlab{a}})Li, Hammoud, Itani, Khizbullin, and
  Ghanem}]{Li2023CAMELCA}
G.~Li, Hasan Abed Al~Kader Hammoud, Hani Itani, Dmitrii Khizbullin, and Bernard
  Ghanem. 2023{\natexlab{a}}.
\newblock Camel: Communicative agents for "mind" exploration of large scale
  language model society.
\newblock \emph{ArXiv}, abs/2303.17760.

\bibitem[{Li et~al.(2023{\natexlab{b}})Li, Wang, Wu, Zhang, Xu, Fu, Tiwari,
  Wan, and Wang}]{li2023huatuo26m}
Jianquan Li, Xidong Wang, Xiangbo Wu, Zhiyi Zhang, Xiaolong Xu, Jie Fu, Prayag
  Tiwari, Xiang Wan, and Benyou Wang. 2023{\natexlab{b}}.
\newblock \href {http://arxiv.org/abs/2305.01526} {Huatuo-26m, a large-scale
  chinese medical qa dataset}.

\bibitem[{Lin(2004)}]{lin2004rouge}
Chin-Yew Lin. 2004.
\newblock \href {https://aclanthology.org/W04-1013} {{ROUGE}: A package for
  automatic evaluation of summaries}.
\newblock In \emph{Text Summarization Branches Out}, pages 74--81, Barcelona,
  Spain. Association for Computational Linguistics.

\bibitem[{Liu et~al.(2023{\natexlab{a}})Liu, Liao, Meng, Wang, and
  Wang}]{MedicalGPT-zh}
Hongcheng Liu, Yusheng Liao, Yutong Meng, Yu~Wang, and Yanfeng Wang.
  2023{\natexlab{a}}.
\newblock Medicalgpt-zh.
\newblock \url{https://github.com/MediaBrain-SJTU/MedicalGPT-zh}.

\bibitem[{Liu et~al.(2023{\natexlab{b}})Liu, Iter, Xu, Wang, Xu, and
  Zhu}]{liu2023gpteval}
Yang Liu, Dan Iter, Yichong Xu, Shuohang Wang, Ruochen Xu, and Chenguang Zhu.
  2023{\natexlab{b}}.
\newblock Gpteval: Nlg evaluation using gpt-4 with better human alignment.
\newblock \emph{arXiv preprint arXiv:2303.16634}.

\bibitem[{Longpre et~al.(2023)Longpre, Hou, Vu, Webson, Chung, Tay, Zhou, Le,
  Zoph, Wei, and Roberts}]{longpre2023flan}
Shayne Longpre, Le~Hou, Tu~Vu, Albert Webson, Hyung~Won Chung, Yi~Tay, Denny
  Zhou, Quoc~V. Le, Barret Zoph, Jason Wei, and Adam Roberts. 2023.
\newblock \href {https://doi.org/10.48550/ARXIV.2301.13688} {The flan
  collection: Designing data and methods for effective instruction tuning}.

\bibitem[{Mishra et~al.(2023)Mishra, Yao, Chen, Wang, Mittal, and
  Yu}]{mishra2023synthetic}
Prakamya Mishra, Zonghai Yao, Shuwei Chen, Beining Wang, Rohan Mittal, and Hong
  Yu. 2023.
\newblock Synthetic imitation edit feedback for factual alignment in clinical
  summarization.
\newblock \emph{arXiv preprint arXiv:2310.20033}.

\bibitem[{OpenAI(2023)}]{openai2023gpt4}
OpenAI. 2023.
\newblock Gpt-4 technical report.
\newblock \emph{arXiv preprint arXiv:2303.08774}.

\bibitem[{Ortega et~al.(2023)Ortega, Hidrue, Lehrhoff, Ellis, Sisodia, Curry,
  Del~Carmen, and Wasfy}]{ortega2023patterns}
Marcus~V Ortega, Michael~K Hidrue, Sara~R Lehrhoff, Dan~B Ellis, Rachel~C
  Sisodia, William~T Curry, Marcela~G Del~Carmen, and Jason~H Wasfy. 2023.
\newblock Patterns in physician burnout in a stable-linked cohort.
\newblock \emph{JAMA Network Open}, 6(10):e2336745--e2336745.

\bibitem[{Panait and Luke(2005)}]{Panait2005CooperativeML}
Liviu Panait and Sean Luke. 2005.
\newblock Cooperative multi-agent learning: The state of the art.
\newblock \emph{Autonomous Agents and Multi-Agent Systems}, 11:387--434.

\bibitem[{Pereira et~al.(2022)Pereira, Pentyala, Nascimento, Sousa~Jr, and
  De~Cock}]{pereira2022secure}
Mayana Pereira, Sikha Pentyala, Anderson Nascimento, Rafael T~de Sousa~Jr, and
  Martine De~Cock. 2022.
\newblock Secure multiparty computation for synthetic data generation from
  distributed data.
\newblock \emph{arXiv preprint arXiv:2210.07332}.

\bibitem[{Podder et~al.(2021)Podder, Lew, and Ghassemzadeh}]{podder2021soap}
V~Podder, V~Lew, and S~Ghassemzadeh. 2021.
\newblock Soap notes.[updated 2021 sep 2].
\newblock \emph{StatPearls [Internet]. StatPearls Publishing. Available from:
  https://www. ncbi. nlm. nih. gov/books/NBK482263}.

\bibitem[{Radev et~al.(2002)Radev, Qi, Wu, and Fan}]{radev2002evaluating}
Dragomir~R Radev, Hong Qi, Harris Wu, and Weiguo Fan. 2002.
\newblock Evaluating web-based question answering systems.
\newblock In \emph{LREC}. Citeseer.

\bibitem[{Ramprasad et~al.(2023)Ramprasad, Ferracane, and
  Selvaraj}]{ramprasad2023generating}
Sanjana Ramprasad, Elisa Ferracane, and Sai~P Selvaraj. 2023.
\newblock Generating more faithful and consistent soap notes using
  attribute-specific parameters.

\bibitem[{Ravaut et~al.(2022)Ravaut, Joty, and Chen}]{ravaut2022summareranker}
Mathieu Ravaut, Shafiq Joty, and Nancy Chen. 2022.
\newblock \href {https://doi.org/10.18653/v1/2022.acl-long.309}
  {{S}umma{R}eranker: A multi-task mixture-of-experts re-ranking framework for
  abstractive summarization}.
\newblock In \emph{Proceedings of the 60th Annual Meeting of the Association
  for Computational Linguistics (Volume 1: Long Papers)}, pages 4504--4524,
  Dublin, Ireland. Association for Computational Linguistics.

\bibitem[{Rindfleisch(1997)}]{rindfleisch1997privacy}
Thomas~C Rindfleisch. 1997.
\newblock Privacy, information technology, and health care.
\newblock \emph{Communications of the ACM}, 40(8):92--100.

\bibitem[{Shafquat et~al.(2022)Shafquat, Mezey, Beigi, Sun, and
  Aptekar}]{shafquat2022source}
Afrah Shafquat, Jason Mezey, Mandis Beigi, Jimeng Sun, and Jacob~W Aptekar.
  2022.
\newblock A source data privacy framework for synthetic clinical trial data.
\newblock In \emph{NeurIPS 2022 Workshop on Synthetic Data for Empowering ML
  Research}.

\bibitem[{Shinn et~al.(2023)Shinn, Cassano, Labash, Gopinath, Narasimhan, and
  Yao}]{shinn2023reflexion}
Noah Shinn, Federico Cassano, Beck Labash, Ashwin Gopinath, Karthik Narasimhan,
  and Shunyu Yao. 2023.
\newblock \href {http://arxiv.org/abs/2303.11366} {Reflexion: Language agents
  with verbal reinforcement learning}.

\bibitem[{Soldaini(2016)}]{Soldaini2016QuickUMLSAF}
Luca Soldaini. 2016.
\newblock Quickumls: a fast, unsupervised approach for medical concept
  extraction.

\bibitem[{Song et~al.(2020)Song, Tian, Wang, and Xia}]{song-summarizing}
Yan Song, Yuanhe Tian, Nan Wang, and Fei Xia. 2020.
\newblock Summarizing medical conversations via identifying important
  utterances.
\newblock In \emph{Proceedings of the 28th International Conference on
  Computational Linguistics}, pages 717--729, Barcelona, Spain (Online).

\bibitem[{Su et~al.(2022)Su, Zhang, Hassanzadeh, and Schaaf}]{Su2022ExtractAA}
Jing Su, Longxiang Zhang, Hamidreza Hassanzadeh, and Thomas Schaaf. 2022.
\newblock Extract and abstract with bart for clinical notes from doctor-patient
  conversations.
\newblock In \emph{Interspeech}.

\bibitem[{Taori et~al.(2023)Taori, Gulrajani, Zhang, Dubois, Li, Guestrin,
  Liang, and Hashimoto}]{alpaca}
Rohan Taori, Ishaan Gulrajani, Tianyi Zhang, Yann Dubois, Xuechen Li, Carlos
  Guestrin, Percy Liang, and Tatsunori~B. Hashimoto. 2023.
\newblock Stanford alpaca: An instruction-following llama model.
\newblock \url{https://github.com/tatsu-lab/stanford_alpaca}.

\bibitem[{Touvron et~al.(2023)Touvron, Lavril, Izacard, Martinet, Lachaux,
  Lacroix, Rozi{\`e}re, Goyal, Hambro, Azhar, Rodriguez, Joulin, Grave, and
  Lample}]{touvron2023llama}
Hugo Touvron, Thibaut Lavril, Gautier Izacard, Xavier Martinet, Marie-Anne
  Lachaux, Timoth{\'e}e Lacroix, Baptiste Rozi{\`e}re, Naman Goyal, Eric
  Hambro, Faisal Azhar, Aurelien Rodriguez, Armand Joulin, Edouard Grave, and
  Guillaume Lample. 2023.
\newblock Llama: Open and efficient foundation language models.
\newblock \emph{arXiv preprint arXiv:2302.13971}.

\bibitem[{Tsichlis et~al.(2021)Tsichlis, Del~Re, and
  Carmody}]{tsichlis2021past}
Jason~T Tsichlis, Andrew~M Del~Re, and J~Bryan Carmody. 2021.
\newblock The past, present, and future of the united states medical licensing
  examination step 2 clinical skills examination.
\newblock \emph{Cureus}, 13(8).

\bibitem[{Valmeekam et~al.(2023)Valmeekam, Marquez, Sreedharan, and
  Kambhampati}]{valmeekam2023planning}
Karthik Valmeekam, Matthew Marquez, Sarath Sreedharan, and Subbarao
  Kambhampati. 2023.
\newblock On the planning abilities of large language models--a critical
  investigation.
\newblock \emph{arXiv preprint arXiv:2305.15771}.

\bibitem[{Wang et~al.(2023)Wang, Liu, Xi, Qiang, Zhao, Qin, and
  Liu}]{wang2023huatuo}
Haochun Wang, Chi Liu, Nuwa Xi, Zewen Qiang, Sendong Zhao, Bing Qin, and Ting
  Liu. 2023.
\newblock Huatuo: Tuning llama model with chinese medical knowledge.
\newblock \emph{arXiv preprint arXiv:2304.06975}.

\bibitem[{Xiong et~al.(2023)Xiong, Wang, Zhu, Zhao, Liu, Wang, and
  Shen}]{xiong2023doctorglm}
Honglin Xiong, Sheng Wang, Yitao Zhu, Zihao Zhao, Yuxiao Liu, Qian Wang, and
  Dinggang Shen. 2023.
\newblock Doctorglm: Fine-tuning your chinese doctor is not a herculean task.
\newblock \emph{arXiv preprint arXiv:2304.01097}.

\bibitem[{Yao et~al.(2023)Yao, Schloss, and Selvaraj}]{yao2023improving}
Zonghai Yao, Benjamin~J Schloss, and Sai~P Selvaraj. 2023.
\newblock Improving summarization with human edits.
\newblock \emph{arXiv preprint arXiv:2310.05857}.

\bibitem[{Yao and Yu(2021)}]{yao2021improving}
Zonghai Yao and Hong Yu. 2021.
\newblock Improving formality style transfer with context-aware rule injection.
\newblock \emph{arXiv preprint arXiv:2106.00210}.

\bibitem[{Yim et~al.(2023)Yim, Fu, {Ben Abacha}, Snider, Lin, and
  Yetisgen}]{aci-demo}
Wen{-}wai Yim, Yujuan Fu, Asma {Ben Abacha}, Neal Snider, Thomas Lin, and
  Meliha Yetisgen. 2023.
\newblock Aci-bench: a novel ambient clinical intelligence dataset for
  benchmarking automatic visit note generation.
\newblock \emph{Submitted to Nature Scientific Data}.

\bibitem[{Yim and Yetisgen-Yildiz(2021)}]{yim2021towards}
Wen-wai Yim and Meliha Yetisgen-Yildiz. 2021.
\newblock Towards automating medical scribing: Clinic visit dialogue2note
  sentence alignment and snippet summarization.
\newblock In \emph{Proceedings of the Second Workshop on Natural Language
  Processing for Medical Conversations}, pages 10--20.

\bibitem[{Yuan et~al.(2023)Yuan, Yang, Liu, Tian, Liang, Xiao, and
  Xie}]{yuan2023causality}
Siyu Yuan, Deqing Yang, Jinxi Liu, Shuyu Tian, Jiaqing Liang, Yanghua Xiao, and
  Rui Xie. 2023.
\newblock Causality-aware concept extraction based on knowledge-guided
  prompting.
\newblock \emph{arXiv preprint arXiv:2305.01876}.

\bibitem[{Yunxiang et~al.(2023)Yunxiang, Zihan, Kai, Ruilong, and
  You}]{yunxiang2023chatdoctor}
Li~Yunxiang, Li~Zihan, Zhang Kai, Dan Ruilong, and Zhang You. 2023.
\newblock Chatdoctor: A medical chat model fine-tuned on llama model using
  medical domain knowledge.
\newblock \emph{arXiv preprint arXiv:2303.14070}.

\bibitem[{Zeng et~al.(2020)Zeng, Yang, Ju, Yang, Wang, Zhang, Zhou, Zeng, Dong,
  Zhang et~al.}]{zeng2020meddialog}
Guangtao Zeng, Wenmian Yang, Zeqian Ju, Yue Yang, Sicheng Wang, Ruisi Zhang,
  Meng Zhou, Jiaqi Zeng, Xiangyu Dong, Ruoyu Zhang, et~al. 2020.
\newblock Meddialog: Large-scale medical dialogue datasets.
\newblock In \emph{Proceedings of the 2020 Conference on Empirical Methods in
  Natural Language Processing (EMNLP)}, pages 9241--9250.

\bibitem[{Zhang et~al.(2023)Zhang, Chen, Jiang, Yu, Chen, Li, Chen, Wu, Zhang,
  Xiao, Wan, Wang, and Li}]{huatuogpt-2023}
Hongbo Zhang, Junying Chen, Feng Jiang, Fei Yu, Zhihong Chen, Jianquan Li,
  Guiming Chen, Xiangbo Wu, Zhiyi Zhang, Qingying Xiao, Xiang Wan, Benyou Wang,
  and Haizhou Li. 2023.
\newblock Huatuogpt, towards taming language models to be a doctor.
\newblock \emph{arXiv preprint arXiv:2305.15075}.

\bibitem[{Zhao et~al.(2023)Zhao, Jin, Chen, Peng, and Yu}]{zhao2023pmcpatients}
Zhengyun Zhao, Qiao Jin, Fangyuan Chen, Tuorui Peng, and Sheng Yu. 2023.
\newblock \href {http://arxiv.org/abs/2202.13876} {Pmc-patients: A large-scale
  dataset of patient summaries and relations for benchmarking retrieval-based
  clinical decision support systems}.

\bibitem[{Zheng et~al.(2023)Zheng, Chiang, Sheng, Zhuang, Wu, Zhuang, Lin, Li,
  Li, Xing, Zhang, Gonzalez, and Stoica}]{zheng2023judging}
Lianmin Zheng, Wei-Lin Chiang, Ying Sheng, Siyuan Zhuang, Zhanghao Wu, Yonghao
  Zhuang, Zi~Lin, Zhuohan Li, Dacheng Li, Eric.~P Xing, Hao Zhang, Joseph~E.
  Gonzalez, and Ion Stoica. 2023.
\newblock \href {http://arxiv.org/abs/2306.05685} {Judging llm-as-a-judge with
  mt-bench and chatbot arena}.

\bibitem[{Zhu et~al.(2018)Zhu, Lu, Zheng, Guo, Zhang, Wang, and
  Yu}]{zhu2018texygen}
Yaoming Zhu, Sidi Lu, Lei Zheng, Jiaxian Guo, Weinan Zhang, Jun Wang, and Yong
  Yu. 2018.
\newblock Texygen: A benchmarking platform for text generation models.
\newblock \emph{SIGIR}.

\end{thebibliography}
\bibliographystyle{acl_natbib}

\newpage

\appendix

\section{Appendix}
\label{sec:appendix}

\subsection{SOAP Structure}
\label{SOAP-Structure-appendix}
The SOAP (Subjective, Objective, Assessment, and Plan) structure is commonly used by providers \cite{podder2021soap}.

\begin{enumerate}[topsep=0.5pt,itemsep=0.2ex,partopsep=0.2ex,parsep=.20ex]
    \item The Subjective section is a detailed report of the patient’s current conditions, such as source, onset, and duration of symptoms, mainly based on the patient’s self-report. This section usually includes chief ccomplaint, history of present illness and symptoms, current medications, and allergies.
    \item The Objective section documents the results of physical exam findings, laboratory data, vital signs, and descriptions of imaging results.
    \item The Assessment section typically contains medical diagnoses and reasons that lead to medical diagnoses. The assessment is typically based on the content from the chief complaint, and the subjective and objective sections.
    \item The Plan section addresses treatment plans based on the assessment.
\end{enumerate}

\subsection{Prompts for ChatGPT\&GPT4}
\label{baseline_prompt}
We use the following prompts to instruct ChatGPT and GPT4 to generate the synthetic patient-physician dialogue based on the provided clinical note.

\emph{Generate the conversation between physician and patient. But for some cases, if the patient eventually dies (according to the clinical note), you can add the patient’s family at the end of the conversation to make it more reasonable. The conversation should include all the information in the following note, especially paying attention to those numbers and medical concepts. The conversation can be more colloquial. When the physician is speaking, the patient can have many modal particles (e.g. hmm, yes, okay) to increase interaction. All the numbers and medical concepts that appear in the note should be mentioned by the physician. Professional medical terms and numbers should more likely occur in the physician’s utterances but not in the patient’s answer. The physician may describe and explain professional judgment to the patient and instruct the patient on follow-up requirements but not ask questions that require professional medical knowledge to answer. The patient’s answer should be succinct and accurate in a colloquial lay language style.}

\subsection{Experimental Settings}

In our study on generating conversation datasets using ChatGPT and GPT-4, we adopted a temperature setting of 0.7. This setting was consistently applied across our methodologies. For each round of dialogue, we set the max tokens for physician role-play as 200 tokens and the patient role-play as 100 tokens. For the intrinsic evaluation phase, we selected a subset of 20 data points from the MT-Dialog dataset and randomly chose 100 datasets from the pmc dataset for testing. In terms of external evaluation, we selected three random data points from each model's output on the pmc dataset to use as few-shot examples. These were inputted into GPT-4, which then generated dialogues from clinical notes or clincal notes from conversations based 20 data sets from the MT-Dialog dataset. During the external chatbot evaluation, we used 10k datasets generated by ChatGPT, GPT-4, and NoteChat-ChatGPT to fine-tune LLaMA2-7b on two A100-40g gpus. During the fine-tuning process, we used DeepSpeed Zero-2 for training, with a learning rate of 2e-5, a batch size of 16, max tokens of 4048 and 1 training epochs. We employ the same settings to train LLaMA2-7b for the generation of clinical tasks from dialogues and the dialogues from clinical notes.

\subsection{Color for Polish Promopt}

We have used consistently different colors to indicate in the polish prompt, as shown in Table~\ref{polish_prompt}, which parts of our prompt have achieved these ten different functions.

\begin{enumerate}[topsep=1.0pt,itemsep=0.5ex,partopsep=0.5ex,parsep=.20ex]
    \item \textcolor{yellow}{Yellow:} Make the conversation as colloquial as possible
    \item \textcolor{Orchid}{Orchid:} Increase the number of rounds of interaction
    \item \textcolor{pink}{Pink:} Professional terms and vocabulary should come from the physicians, and patients should be more colloquial
    \item \textcolor{gray}{Gray:} Basic symptoms and medical history should come from the patient, not the physician
    \item \textcolor{BrickRed}{BrickRed: }The questions asked by the physician should be around the case (to avoid hallucination)
    \item \textcolor{SkyBlue}{SkyBlue:} Physician inquiries should be logical
    \item \textcolor{Emerald}{Emerald: }If there are multiple consultation records, you can split a conversation into multiple ones and then link them with transfer words (e.g., a few days later)
    \item \textcolor{BurntOrange}{BurntOrange: }Range of rounds of interaction
    \item \textcolor{Thistle}{Thistle: }Must contain the given keywords
    \item \textcolor{Periwinkle}{Periwinkle: }Do not generate duplicate information
\end{enumerate}

Note that there are some similar and repeated parts in the prompt, which are because we found that mentioning a certain point multiple times in different places in the prompt is more helpful for LLM to avoid certain problems.

\begin{table}[h]
\centering
\scalebox{0.7}{
\begin{tabular}{cccc}

\hline

\textbf{ Group } & \textbf{ Our Score } & \textbf{ GPT-4 Score } & \textbf{ ChatGPT Score } \\
\hline physicians & 0.78 &	0.80 &	0.93 \\
 Crowd &	0.70 &	0.75 &	0.90 \\
\hline
\end{tabular}
}
\caption{To evaluate the annotation consistency of the annotators, we calculated the agreement score (Cohen's kappa coefficient) for both the expert group and the crowd group. For each group, we calculated the agreement score for the annotators ranking NoteChat, GPT-4, and ChatGPT as the first, to determine whether the annotators consistently labeled the same model as the best.}
\vspace{-1mm}
\end{table}

\begin{table}[h]
\centering
\scalebox{0.9}{
\begin{tabular}{cc}

\hline

\textbf{ Comparison } & \textbf{ Win Rate } \\
\hline NoteChat-GPT-4 -> our & 0.7 \\
 NoteChat-GPT-4 -> GPT-4 &	0.7 \\
NoteChat-GPT-4 -> ChatGPT &	1.0 \\
\hline
\end{tabular}
}
\caption{To empirically validate the superiority of our approach over GPT-4, we employed the NoteChat-GPT4 version to demonstrate that our model consistently outperforms GPT-4. After replacing the gpt3.5-turbo module in NoteChat model with GPT-4, we generated a new set of dialogues and compared them with NoteChat-GPT3, GPT4, and ChatGPT respectively. For each comparison, we asked GPT-4 to judge and choose the best dialogue. For the same dialogue comparison between different models, we changed the order to avoid the order influencing GPT-4's judgment. Finally, we obtained the win rate as shown in the experimental results:}
\vspace{-1mm}
\end{table}

\begin{table}[h]
\setlength{\tabcolsep}{3.9pt}
\renewcommand{\arraystretch}{1.05}
\centering
\begin{footnotesize}
\begin{tabularx}{0.48\textwidth}{X} 

\toprule
In this task, we ask for your expertise in annotating the quality of system-generated replies by machine learning models. Mainly we provide the history dialogue along with system-generated replies and ask for your preference.
\newline
\newline
Output your ranking for system-generated replies. Use the following format, and do not add any other text.
\newline
\newline
Some examples:
\newline
$a>b>c>d>e$
\newline
$e>d>c>b>a$
\newline
\newline
History Conversation:
\newline
[\emph{History Conversation}]
\newline
\newline
Conversation snippet:
\newline
[\emph{utterance}]
\newline
\newline
\newline
\newline
System-generated summaries:
\newline
1. [\emph{Utterance1}]
\newline
2. [\emph{Utterance2}]
\newline
3. [\emph{Utterance3}]
\newline
4. [\emph{Utterance4}]
\newline
5. [\emph{Utterance5}]
\newline
\newline
Now, output your ranking:
\\
\bottomrule
\end{tabularx}
\end{footnotesize}
\vspace{-0.05in}
\caption{GPT-4 Prompt for preference ranking in extrinsic evaluation.}
\label{extrinsic_eval_prompt}
\vspace{-0.15in}
\end{table}

\begin{table}[h]
\setlength{\tabcolsep}{3.9pt}
\renewcommand{\arraystretch}{1.05}
\centering
\begin{footnotesize}
\begin{tabularx}{0.48\textwidth}{X} 

\toprule
In this task, we ask for your expertise in annotating the quality of the system-generated dialogues by machine learning models. Mainly we provide the ground truth dialogue and the clinical note along with system-generated dialogues and ask for your preference.
\newline
\newline
Output your ranking for system-generated dialogues. Use the following format, and do not add any other text.
\newline
\newline
Some examples:
\newline
$a>b>c$
\newline
$c>b>a$
\newline
\newline
Clinical Note:
\newline
[\emph{Clinical Note}]
\newline
\newline
Ground Truth Dialogue:
\newline
[\emph{dialogue}]
\newline
\newline
\newline
\newline
System-generated summaries:
\newline
1. [\emph{dialogue1}]
\newline
2. [\emph{dialogue2}]
\newline
3. [\emph{dialogue3}]
\newline
\newline
\newline
Now, output your ranking:
\\
\bottomrule
\end{tabularx}
\end{footnotesize}
\vspace{-0.05in}
\caption{GPT-4 Prompt for preference ranking in human evaluation.}
\label{human_eval_prompt}
\vspace{-0.15in}
\end{table}




\begin{table*}
    \centering
    \begin{tabularx}{\textwidth}{p{2.5cm}p{5cm}p{7.2cm}}
    \hline
    \footnotesize{Section} & \footnotesize{Subsection} & \footnotesize{Definition}
    \\
    \hline
    \footnotesize{Subjective} & &
    \\

    & \footnotesize{Chief Complaint} 
    & \footnotesize{Patient’s primary motivation for the visit and type of visit}
    \\
    & \footnotesize{Review of Systems}		
    & \footnotesize{Patient’s report of system related health and symptoms}
    \\
    & \footnotesize{Past Medical History}
    & \footnotesize{Patient’s reported diagnoses/conditions (when and what,  excluding laboratory and imaging results and surgeries)}
    \\
    & \footnotesize{Past Surgical History}
    & \footnotesize{Patient’s reported prior surgeries (what, when, where)}
    \\
    & \footnotesize{Family Medical History}
    & \footnotesize{Conditions affecting patient’s close genetic relatives}
    \\
    & \footnotesize{Social History}
    & \footnotesize{Patient’s alcohol, tobacco, and drug related behaviors}
    \\
    & \footnotesize{Medications}
    & \footnotesize{Patient’s list of medications (not prescribed during visit)}
    \\
    & \footnotesize{Allergies}
    & \footnotesize{Patient’s list of allergies (primarily medicinal)}
    \\
    & \footnotesize{Miscellaneous}
    & \footnotesize{Patient’s clinically relevant social and other circumstances}
    \\
    \hline
    \footnotesize{Objective} & &
    \\
    & \footnotesize{Immunizations}
    & \footnotesize{Vaccination record (not frequently discussed)}
    \\
    & \footnotesize{Laboratory and Imaging Results}
    & \footnotesize{Clinician’s discussion of laboratory/imaging results}
    \\
    \hline

    \footnotesize{Assessment} & &
    \\
    & \footnotesize{Assessment}
    & \footnotesize{Synthesis of reason for visit and pertinent diagnosis}
    \\
    \hline
    \footnotesize{Plan} & &
    \\
    & \footnotesize{Diagnostics \& Appointments}
    & \footnotesize{Plan for future tests, appointments, or surgeries}
    \\
    & \footnotesize{Prescriptions \& Therapeutics}
    & \footnotesize{Plan for medications and therapeutics}
    \\

    \hline
    \end{tabularx}
    \caption{Details of the SOAP structure.} 
    \label{table:SOAP-structure-appendix}
    \vspace{-5mm}
\end{table*}




\begin{table*}[hbt]

\scalebox{0.75}{
\begin{tabular}{ll}
\hline
\multirow{18}{*}{\textbf{Planning Module}} 
& Apply the physician and Patient prompt to generate the beginning and lead the physician LLM to ask about the\\& medical record. Continue to generate 20 to 40 utterances conversations between physician and patient to ask  \\ & or tell the patient regarding the case(you must follow up the history conversation). The conversations you \\&generate must cover all the keywords I gave you. You cannot revise or eliminate any keywords and\\& you cannot use synonyms of the keywords. Your conversation should also include all information. \\&If it's difficult to include all the information and key words, you can use the \\&original sentences in the clinical note.\\
& The Clinical Note: \textcolor{blue}{Clinical Note}\\
& The Key Words: \textcolor{red}{$key_1$, $key_2$,...}\\
&Your conversations must include all the keywords I provided to you, and if it's not possible to\\ &include them all, you can make slight modifications based on the original wording in the notes. \\ &You cannot revise or eliminate any key words and you cannot use synonyms of the keywords. \\ & Your conversation should also include all information. If it's difficult to include all the information\\ & and key words, you can use the original sentences in the clinical note. Your generation must \\ &follow the logical sequence of a physician's inquiry. Your conversations must follow the logical \\& sequence of a physician's inquiry. For example, the general logical order of the conversation is: first \\ & discussing symptoms, then discussing the medical history, followed by discussing testing and \\& results, and finally discussing the conclusion and treatment options, etc. The physician didn't know \\& any information of medical history or symptoms. This information should be told by the patient\\

\hline
\end{tabular}
}
\caption{Planning Module prompt.}
\label{planning_prompt}
\end{table*}

\begin{table*}[hbt]

\scalebox{0.75}{
\begin{tabular}{ll}
\hline

\multirow{26}{*}{\textbf{Physician Prompt}} 
& Please role-play as a physician and further generate questions or conclusion, or the test \\ &result(such as medication test result or vital signs) based on the above dialogue and clinical\\ &note(after mentioned examination, you have to know test results and vital signs so you shouldn't ask \\&the patient about a test result or vital signs). Add 'physician:' before each round. Your question, \\&answer or conclusion(tell the patient the test result) should be around the keywords (I gave you)\\ &corresponding to the clinical note(finally, the whole conversation should include all the keywords).\\ & the answer of your questions can be found on the clinical note. You cannot modify these key\\ &words or use synonyms. You need to ensure the treatment plan, medication, and dosage you give to\\ &the patient must also be totally consistent with the clinical note. Do not ask questions which\\ &answers cannot be found in the clinical note. You may describe and explain professional judgment to\\ &the patient and instruct the patient on follow-up requirements, but not ask questions that require\\ &professional medical knowledge to answer. The order of the questions you ask must match the order\\ &of the keywords I provided. If it's not possible to include them all, you can make slight modifications\\ &based on the original wording in the notes. If the history conversation has included\\ &the keywords, there is no need to include them again. The treatment plan and conclusions\\ &you provide must align completely with the clinical notes. Do not add treatment plans\\ &that is not present in the clinical notes. You don't know the patient's medical history and symptoms.\\& You should ask or lead the patient to tell you the symptoms and his medical history, and you\\& don't have any information about his medical history and symptoms. All the information of medical\\ &history, symptoms, medication history, and vaccination history should be told by the patient. You can\\ &tell the patient the test results, vital signs, and some conclusions.\\
& The Clinical Note: \textcolor{blue}{Clinical Note}\\
& The Key Words: \textcolor{red}{$key_1$, $key_2$,...}\\
& The History Conversation: \textcolor{purple}{History Dialogue}\\
&You should only generate one utterance based on history conversation. Remember, you are the physician, not the patient.\\ & Don't mention the information that has been mentioned in history conversation. If you feel that the patient's\\ &information is incomplete, you can supplement it based on the clinical note and include relevant\\ &keywords. However, please refrain from saying, 'based on medical record or clinical note.'\\& Instead, you should say, 'I guess...'\\

\hline

\hline
\multirow{16}{*}{\textbf{Patient Prompt}} 
& Act as a patient to reply to the physician. Add 'Patient:' before each round. Your answer should \\& align with the clinical notes. You are just an ordinary person. Your response should be made as\\ &colloquial as possible. Don't mention any experimental results, conclusions, or medical dosage.\\ &because you're just an ordinary person and may not understand the meaning of these results. \\ & But you could tell the physician your medical history, medication history, or vaccination history\\&(medical history, medication history, or vaccination history are all long to medical history). \\& Your response should revolve around the physician's words and avoid adding information that was not mentioned.\\
& The Clinical Note: \textcolor{blue}{Clinical Note}\\
& The History Conversation: \textcolor{purple}{History Dialogue}\\
& Your reply should be succinct and accurate in a colloquial lay language style and must be aligned\\ & with clinical notes. Don't generate the part which should be said by the physician. Do not say all the\\ &information unless the physician asks about it. You cannot say any information about your test result\\ & or vital signs. Your medical history, vaccination history, and medication history all belong to\\ & medical history. Your reply must be completely aligned with the clinical note. But you cannot say any\\ & examination or test results because you are not a physician. You must not be able to use highly\\ & specialized terms or medical terminology. You can only describe limited common symptoms. \\& You shouldn't use the abbreviation if you know the full name(you should use the full name, not the abbreviation,\\ & such as D9 must be day 9, D7 must be day 7\\
\hline

\end{tabular}
}
\caption{Roleplay module prompt for physician role and patient role.}
\label{roleplay_prompt}
\end{table*}

\begin{table*}[hbt]

\scalebox{0.75}{
\begin{tabular}{ll}

\hline

\multirow{61}{*}{\textbf{Polish Prompt}} 
& \colorbox{Orchid}{Expand the conversation.} \colorbox{yellow}{The conversation for patient parts can be more colloquial.} When the physician\\ & is speaking, \colorbox{green}{the patient can have many modal particles (e.g. hmm, yes, okay) to increase interaction.}\\
& \colorbox{pink}{All the numbers and medical concepts that appear in the note should be mentioned by the physician.}\\
& \colorbox{pink}{Professional medical terms and numbers should always occur in the physician's utterances but not in}\\ & \colorbox{pink}{the patient's answer.} The physician may describe and explain professional judgment to the patient\\ & and instruct the patient on follow-up requirements, \colorbox{pink}{but not ask questions that require professional}\\ & \colorbox{pink}{medical knowledge to answer} \colorbox{BrickRed}{and the question must be around the clinical note(the patient could} \\ &\colorbox{BrickRed}{find the answer on the clinical note).} \colorbox{Gray}{All the information of medical history, symptoms and medication} \\ &\colorbox{Gray}{history should be told by patient.} \colorbox{yellow}{The patient's answer should be succinct and accurate in a}\\ & \colorbox{yellow}{colloquial lay language style. The answer should align with the clinical notes and as colloquial}\\&  \colorbox{yellow}{as possible.} \colorbox{Emerald}{You can add some transitional phrases to make the conversation more logical.} \\
& For example:\\
& Example 1:\\
& Patient: I understand, please go ahead.\\
& (After examination)\\
& physician: The result shows....\\
& Example 2:\\
& Patient: Thank you for the diagnosis, physician.\\
& (After two years)\\
& physician: Hi...\\
& Example 3:\\
& Patient: Okay, I understand. \\
& (Few days latter)\\
& physician: Hi...\\
& \colorbox{SkyBlue}{Your conversations must follow the logical sequence of a physician's inquiry. For example, the general}\\ & \colorbox{SkyBlue}{logical order of the conversation is: first discussing symptoms, then discussing the} \\& \colorbox{SkyBlue}{medical history, followed by discussing testing and results, and finally discussing treatment} \\ & \colorbox{SkyBlue}{options, conclusioin etc."} \colorbox{Emerald}{If you find this conversation to be incoherent, you can try dividing it}\\ & \colorbox{Emerald}{into two separate coherent conversations.} \colorbox{Orchid}{Patients should not say too much information at once.}\\
& The Clinical Note: \textcolor{blue}{Clinical Note}\\
& The Key Words: \textcolor{red}{$key_1$, $key_2$,...}\\
& The History Conversation: \textcolor{purple}{Conversation}\\
& There are only one patient and one physician and just return the conversation. You conversation must\\ & include all the key words I gave you. \\
& Your conversation should also include all information. \colorbox{Thistle}{if it's difficult to include them all, you}\\ & \colorbox{Thistle}{can use the original sentences in the notes.} \\
& \colorbox{gray}{The common symptoms and common medical history should be told by the patient.} \\
& \colorbox{pink}{Some specific symptoms and medical history should be added by the physician} \colorbox{gray}{after the patient has}\\ & \colorbox{gray}{finished describing his symptoms and medical history.}\\
& For example:\\
& physician: Can you give me your medical history record?\\
& Patient: Here you are.\\
& physician: Based on your medical history record...\\
& Because after the patient has finished describing common symptoms or medical history, he will give\\ & physician his medical history records. \\
& After patient gives the physician his medical history record, the physician could know medical\\ & history record. Otherwise he didn't know any information of the medical history.\\
& Some results should not come from history clinical note they should come from the examination.\\
& All the examination results, history examination results, vital sigh and medical number must be told by physician.\\
& \colorbox{BurntOrange}{The revised conversation should be at least around 30 to 40 utterances}\\&(the physician or patient should say too much information at once).\\
& \colorbox{Thistle}{The conversation must include all the information on the clinical note.}\\
& \colorbox{Thistle}{You must include all the key words I gave you. If it is difficult to include all the key words you}\\&  \colorbox{Thistle}{could use original the sentences of clinical note.} \\
& \colorbox{Thistle}{You cannot revise or eliminate any key words and you cannot use synonyms of the key words.} \\
& You shouldn't use the abbreviation if you know the full name(you should use full name not \\ & abbreviation, such as D9 must be day 9, D7 must be day 7. If both the full name and the abbreviation \\& appear, it's better to use the full name rather than the abbreviation.\\
& \colorbox{pink}{Patients must not say any highly specialized terms, medical terminology or medical dosage.}\\
& \colorbox{gray}{They can only describe limited common symptoms.}\\& The physician should supplement the remaining information based on test results.\\
& \colorbox{Periwinkle}{Don't repeat the same information in long paragraphs. The utterance of the dialogue needs to be}\\ & \colorbox{Periwinkle}{expanded as much as possible.}\\
\hline
\end{tabular}
}
\caption{Polish prompt.}
\label{polish_prompt}
\end{table*}

\begin{table*}[hbt]

\scalebox{0.75}{
\begin{tabular}{ll}
\hline

\multirow{19}{*}{\textbf{Combine Prompt}} & The above two paragraphs were extracted from a complete conversation. \\& Please concatenate the two dialogues together. Add 'physician:' before the physician's words \\& and 'Patient:' before the patient's words for easier differentiation. \\& Please combine these two dialogues.  \\& It means that your generation should include all the information \\& such as dosage of the medication  which is mentioned in the clinical note \\&if the dosage is not mentioned in the clinical not \\& you should not mention it and the length should be longer than \\& both of these two conversations  even longer than the sum of them. \\&You should try to ensure that the dialogue is smooth, \\& and don't use any greetings such as 'Hi there', 'how are you feeling today?', \\& 'Hey', 'Hello' or any farewells in the dialogue. \\& The entire conversation takes place at the same time and place, \\& and revolves around the same patient and physician. \\& Try to make the conversation smoother. Try to make these two dialogues into one dialogue \\& that takes place at the same time and place. Modify this conversation \\& by deleting all greeting sentences \\&such as 'Hi', 'Hey', 'Hi there', 'How are you feeling today', and 'Good Morning'.\\&  The conversation must include these key words:\textcolor{red}{$key_{1},key_{2},...$} \\& and you should also eliminate the repeat parts.\\
\hline
\end{tabular}
}
\caption{Combine prompt.}
\label{combine_prompt}
\end{table*}

\subsection{Ablation Study for Planning Module}

To demonstrate the importance of the planning module, we designed the following experiment:

We conducted evaluations using both GPT-4 and human assessments. In the absence of the checklist and planning module, relying solely on role play and polishing for dialogue generation, the results were as follows:

\begin{itemize}
    \item \textbf{GPT-4 Evaluation Win Rate:}
    \begin{itemize}
        \item Our model (without checklist \& planning): 32\%
        \item GPT-4: 68\%
    \end{itemize}
    \item \textbf{Human Evaluation Win Rate:}
    \begin{itemize}
        \item Our model (without checklist \& planning): 38\%
        \item GPT-4: 62\%
    \end{itemize}
\end{itemize}

The absence of the checklist and planning module resulted in the model's inability to ensure comprehensive coverage of necessary information. While the generated dialogues were logically coherent, they significantly lacked informational content. This deficiency is primarily attributable to our model being based on GPT-3.5, which has a substantially lower capacity for information coverage compared to GPT-4.

Furthermore, when relying solely on a randomly ordered checklist, the results were as follows:

\begin{itemize}
    \item \textbf{GPT-4 Evaluation Win Rate:}
    \begin{itemize}
        \item Our model (without planning module): 54\%
        \item GPT-4: 46\%
    \end{itemize}
    \item \textbf{Human Evaluation Win Rate:}
    \begin{itemize}
        \item Our model (without planning module): 40\%
        \item GPT-4: 60\%
    \end{itemize}
\end{itemize}

These results indicate slight differences. When evaluated by GPT-4, our model without the planning module appeared superior due to providing more information in shorter dialogue turns and extended conversations. However, human evaluators found the generated dialogues logically disorganized, primarily due to the absence of the planning module. The randomly ordered checklist led to each conversational turn lacking logical progression, making it seem less like a real dialogue. This highlights the critical importance of the planning module.

\end{document}